\documentclass[sigconf]{acmart}
\AtBeginDocument{%
  }

\setcopyright{acmlicensed}

\copyrightyear{2026}
\acmYear{2026}
\setcopyright{cc}
\setcctype{by}
\acmConference[WWW '26]{Proceedings of the ACM Web Conference 2026}{April 13--17, 2026}{Dubai, United Arab Emirates}
\acmBooktitle{Proceedings of the ACM Web Conference 2026 (WWW '26), April 13--17, 2026, Dubai, United Arab Emirates}
\acmPrice{}
\acmDOI{10.1145/3774904.3792262}
\acmISBN{979-8-4007-2307-0/2026/04}

\usepackage{amsmath,amsfonts,bm}

\usepackage{natbib}
\usepackage{url}
\usepackage{hyperref}
\usepackage{enumitem}
\usepackage{float}
\usepackage{multirow}
\usepackage{textcomp}
\usepackage{tabularx}
\usepackage[normalem]{ulem}
\usepackage{footnote}
\usepackage{threeparttable}
\usepackage{caption}
\usepackage[utf8]{inputenc}

\usepackage{amssymb}             
\usepackage{mathrsfs} 
\usepackage{CJKutf8}
\usepackage[table]{xcolor}
\usepackage{indentfirst}
\usepackage{threeparttable}
\usepackage{array}
\usepackage{stfloats}
\usepackage{verbatim}
\usepackage{subfigure}
\usepackage{colortbl}
\usepackage{cleveref}
\usepackage{makecell}
\usepackage{rotating}
\usepackage{arydshln}
\usepackage{wrapfig}
\usepackage{pifont}
\usepackage[colorinlistoftodos]{todonotes}

\usepackage{algorithm}
\usepackage{algpseudocode}

\usepackage[table]{xcolor} 
\usepackage{ulem} 
\usepackage{tikz} 

\definecolor{mlb}{RGB}{173,216,230}  
\definecolor{mlo}{RGB}{255,223,186}  

\begin{document}

\title{Hermes the Polyglot: A Unified Framework to Enhance Expressiveness for Multimodal Interlingual Subtitling}



\author{Chaoqun Cui}
\affiliation{
  \institution{MAIS, Institute of Automation, Chinese Academy of Sciences}
  \city{Beijing}
  \country{China}
}
\affiliation{
  \institution{School of AI, University of\\Chinese Academy of Sciences}
  \city{Beijing}
  \country{China}
}
\email{cuichaoqun2025@ia.ac.cn}

\author{Shijing Wang}
\affiliation{
  \institution{Beijing Jiaotong University}
  \city{Beijing}
  \country{China}
}
\affiliation{
  \institution{Hujing Digital Media \& Entertainment Group}
  \city{Beijing}
  \country{China}
}
\email{shijingwang@bjtu.edu.cn}

\author{Liangbin Huang}
\affiliation{
  \institution{Hujing Digital Media \& Entertainment Group}
  \city{Beijing}
  \country{China}
}
\email{huangliangbin.hlb@alibaba-inc.com}

\author{Qingqing Gu}
\affiliation{
  \institution{Geely AI lab}
  \city{Ningbo}
  \state{Zhejiang}
  \country{China}
}
\email{Qingqing.gu3@geely.com}

\author{Zhaolong Huang}
\affiliation{
  \institution{Hujing Digital Media \& Entertainment Group}
  \city{Beijing}
  \country{China}
}
\email{zhaolong.hzl@alibaba-inc.com}

\author{Xiao Zeng}
\affiliation{
  \institution{Hujing Digital Media \& Entertainment Group}
  \city{Beijing}
  \country{China}
}
\email{zengxiao@alibaba-inc.com}

\author{Wenji Mao}
\authornote{Corresponding author.}
\affiliation{
  \institution{MAIS, Institute of Automation, Chinese Academy of Sciences}
  \city{Beijing}
  \country{China}
}
\affiliation{
  \institution{School of AI, University of\\Chinese Academy of Sciences}
  \city{Beijing}
  \country{China}
}
\email{wenji.mao@ia.ac.cn}







\renewcommand{\shortauthors}{Chaoqun Cui et al.}

\begin{abstract}

Interlingual subtitling, which translates subtitles of visual media into a target language, is essential for entertainment localization but has not yet been explored in machine translation. Although Large Language Models (LLMs) have significantly advanced the general capabilities of machine translation, the distinctive characteristics of subtitle texts pose persistent challenges in interlingual subtitling, particularly regarding semantic coherence, pronoun and terminology translation, and translation expressiveness. To address these issues, we present Hermes, an LLM-based automated subtitling framework. Hermes integrates three modules: Speaker Diarization, Terminology Identification, and Expressiveness Enhancement, which effectively tackle the above challenges. Experiments demonstrate that Hermes achieves state-of-the-art diarization performance and generates expressive, contextually coherent translations, thereby advancing research in interlingual subtitling.

\end{abstract}




\maketitle

\section{Introduction}

The interlingual subtitling task aims to translate the subtitles of visual media programs such as films or TV series into the target language, serving as a crucial component of online entertainment program localization \cite{vd1,vd2,vd3}, while also being a relatively underexplored subfield in machine translation. In recent years, with the rise of streaming and online video platforms such as Netflix and Disney+, along with the globalization and diversification of the visual media industry, the demand for advanced interlingual subtitling technologies has grown substantially. However, despite the rapid development of Large Language Models (LLMs), which has significantly improved the general capabilities of machine translation \cite{sr,ladder}, they reveal clear limitations when applied to in-depth tasks like subtitle translation, including insufficient professional terminology translation \cite{mtchallenge1,mtchallenge2}, deviations from industry standard expressions \cite{mtchallenge4,mtchallenge3}, and weak style adaptability \cite{mtchallenge4,mtchallenge5}.

The characteristics of subtitles are: 1) subtitles are composed of dialogues between characters in a program; 2) subtitle lines are short texts with strong contextual connections; 3) subtitle texts are closely tied to video and speech. These features pose specific challenges for subtitle translation: 1) \textit{semantic coherence}: the limited input token capacity of LLMs prevents feeding an entire subtitle at once, and how multiple lines are segmented directly affects the integrity of LLM’s input boundary and the translation coherence; 2) \textit{pronoun reference}: due to cultural differences, different languages may have distinct pronoun systems (e.g., honorifics in Japanese and Korean, or omission of pronouns in some languages); 3) \textit{terminology translation}: programs contain numerous specialized terms in visual media, requiring accurate and consistent translation; 4) \textit{translation quality}: besides accuracy, audiences prefer expressive translations.

In this study, to address these challenges, we present Hermes, an LLM-based interlingual subtitling framework. Hermes integrates three modules: 1) \textit{Speaker Diarization (SD)}: combining visual and speech modalities to annotate speakers (characters) in visual media scenarios, thereby addressing line segmentation and pronoun translation; 2) \textit{Terminology Identification (TI)}: employing LLM knowledge distillation to identify and translate proper nouns in subtitles; 3) \textit{Expressiveness Enhancement (EE)}: proposing \textbf{S}egment-wise \textbf{A}daptive \textbf{P}reference \textbf{O}ptimization (SAPO) method with LLM-as-a-Judge to enhance the translation expressiveness. Expressiveness refers to that, on the premise of conveying the original meaning (accuracy), the translation should also conform to the grammatical and cultural conventions of the target language (naturalness), and be able to convey stylistic information such as the emotion, tone, and atmosphere of the original text (vividness) \cite{avt2,avt3}.

In summary, the contributions of this study are as follows:
\begin{itemize}
\item We define interlingual subtitling as a multimodal machine translation task and systematically identify the unique challenges this task faces in practical applications.
\item We propose Hermes, an LLM-based subtitling framework supported by prior information from visual and audio modalities and specific preference optimization technique.
\item We construct a multidimensional evaluation framework based on LLM-as-a-Judge for interlingual subtitling.
\item Experimental results demonstrate the excellent performance of Hermes on the interlingual subtitling task.
\end{itemize}

\section{Task Definition and Notations}

The interlingual subtitling task aims to translate subtitles of visual media programs from the source language into the target language, playing a crucial role in making multimedia content more searchable, comprehensible, and shareable in the global web environment. It takes video, speech, and subtitle lines as input and outputs subtitle lines in the target language. A bilingual dataset $\mathbb{D}$ for subtitle translation encompasses source language line set $\mathcal{L}_{\text{src}}$ and target language line set $\mathcal{L}_{\text{tgt}}$ from multiple programs, expressed as $\mathbb{D}\equiv \{(s, t)\in \mathcal{L}_{\text{src}}\times \mathcal{L}_{\text{tgt}}\}$. We collect and release multilingual parallel subtitle corpora from the online video platform Youku. See Appendix~\ref{sec:dc} for details of this dataset. 

Hermes leverages the dataset $\mathbb{D}$ to train a subtitle translation LLM $\pi _{\text{st}}$. The input prompt $x$ to $\pi _{\text{st}}$ comprises a context $\mathcal{C}$ and a set of $n$ source lines $\{s_i \mid i=1,\cdots ,n\}$ to be translated, formally expressed as $x\equiv \mathcal{C}\oplus \{s_i\}$. Beyond introduction and instruction text, $\mathcal{C}$ also includes prescribed term translations and character feature descriptions, which are the outputs of SD and TI modules. The response $y$ generated by $\pi _{\text{st}}$ consists of translations corresponding to each line, represented as $\{t_i\mid i=1,\cdots ,n\}$, i.e., $y\equiv \{t_i\}$. Detailed examples of prompt and response are provided in our \href{https://github.com/CcQunResearch/Hermes/blob/main/Reproducibility%20Supplementary%20Material.pdf}{Supplementary Material} (2.1).

\section{Proposed Framework}

\begin{figure*}[t]
  \centering
  \includegraphics[width=0.95\textwidth]{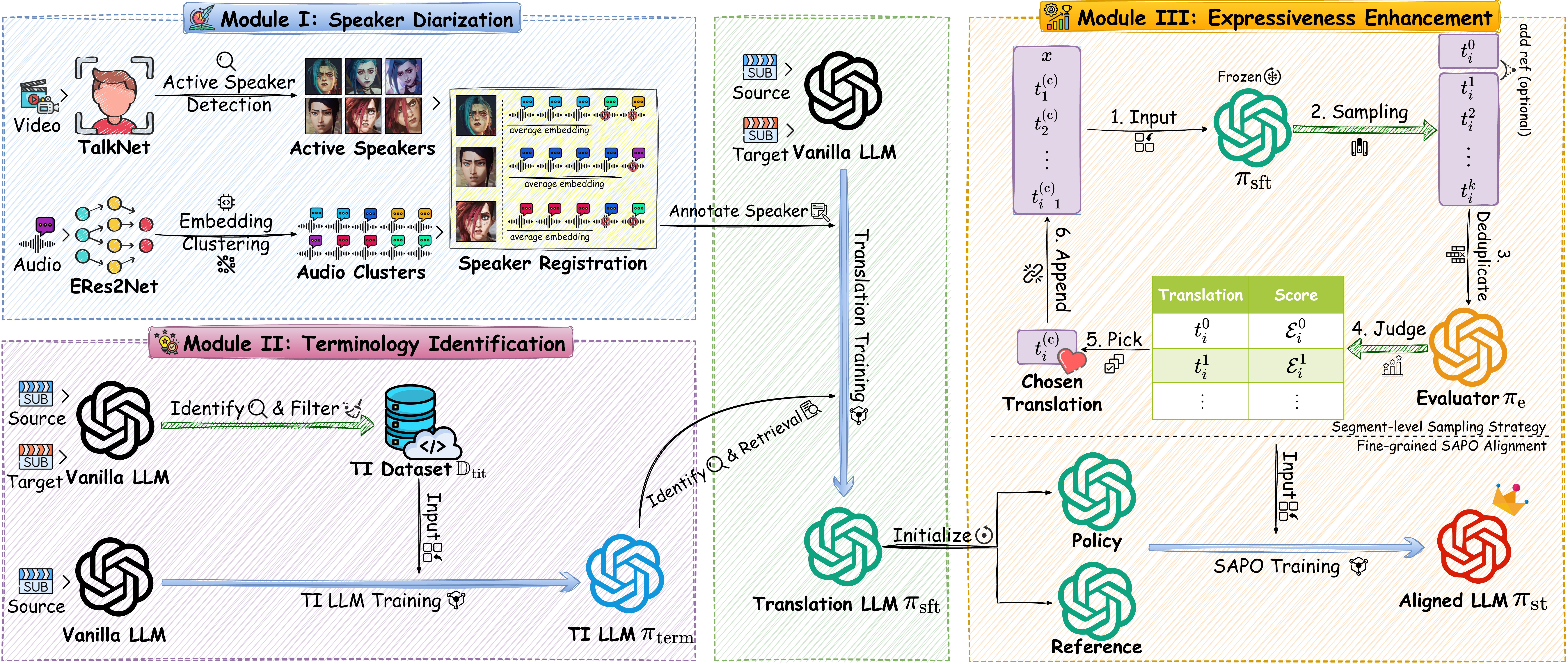}
  \caption{The overall structure of the unified framework Hermes to enhance expressiveness for interlingual subtitling.}
  \label{fig:framework}
\end{figure*}

We present the overall design of Hermes in Figure~\ref{fig:framework}. Speaker Diarization and Terminology Identification modules provide essential prior information. Subsequently, in Expressiveness Enhancement module, we employ SAPO technique to enhance the expressiveness of the translations, thereby obtaining the final translation LLM $\pi _{\text{st}}$.

\subsection{Speaker Diarization}

\subsubsection{Overview}

In most genres of visual media programs such as film and TV series, subtitles consist of dialogues between multiple characters, and annotating the speaker for each line helps improve the accuracy of translating personal pronouns. Moreover, lines are closely contextually connected \cite{tmmspk1,tmmspk2}, so when performing subtitle translation, it is preferable to feed LLMs with consecutive lines rather than translating them sentence by sentence. However, given that the input length of LLMs is usually limited, it is challenging to feed the complete subtitles of a program into the LLM all at once; thus, properly segmenting the lines is crucial to maintaining contextual completeness in LLM input. Through SD module of Hermes, we can: (1) improve the accuracy of personal pronouns; and (2) segment lines while ensuring contextual integrity and coherence.

The visual modality can leverage high-accuracy face-related technologies but faces the audio-visual asynchrony problem, where a portion of lines (about 35\% in our test set) are delivered while the speaker is not visible on screen. In contrast, the speech modality does not suffer from this issue, but speech technologies generally have lower accuracy compared to face-related techniques. Moreover, in existing studies, speaker diarization \cite{ava,aclsd} typically assumes predefined speaker information, whereas in our scenario the exact number of speakers is uncertain. Therefore, SD module aims to integrate both visual and speech modalities of the lines to perform speaker registration and annotation.

\subsubsection{Feature Clustering}

For visual modality, we utilize subtitle timing notes in dataset $\mathbb{D}$ for a given line $s \in \mathcal{L}_{\text{src}}$ (example subtitle format in Table~\ref{tab:subtitle}) to extract its corresponding video segment. We employ TalkNet \cite{talknet} to detect and extract the active speaker from the video frames. Lines for which no active speaker is detected are reserved for subsequent processing. For lines where an active speaker is detected, we extract the face embedding $e_{\text{v}}(s)$ (produced by CurricularFace model \cite{curricular}) and perform clustering (e.g., spectral clustering \cite{sc}) to assign each line a visual cluster label $c_{\text{v}}(s)\in \left \{1,\cdots ,n_{\text{v}}\right \}$, where $n_{\text{v}}$ denotes the number of visual clusters. For the speech modality, we similarly use the subtitle timing notes to retrieve the audio segment for each line $s$. We then apply ERes2NetV2 \cite{eres2netv2} to extract a timbre embedding $e_{\text{a}}(s)$ for each audio. Clustering is also applied to the embeddings of all lines’ speech, thereby assigning each line an audio cluster label $c_{\text{a}}(s)\in \left \{1,\cdots ,n_{\text{a}}\right \}$, where $n_{\text{a}}$ denotes the number of audio clusters.

\subsubsection{Speaker Registration}

\begin{figure}[t]
  \centering
  \includegraphics[width=0.49\textwidth]{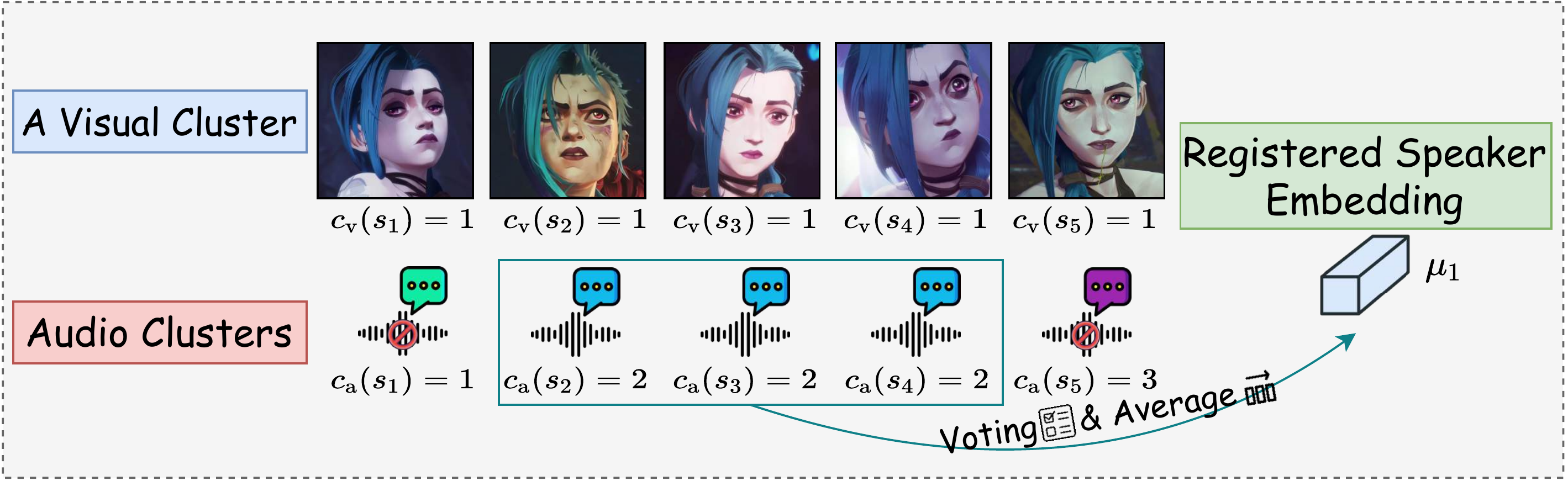}
  \caption{Speaker registration.}
  \label{fig:register}
\end{figure}

Since the facial features extracted from visual modality are more distinctive, its visual clustering results are generally more reliable than those of the speech modality. Therefore, we use visual clustering as the anchor for speaker registration. The speaker registration process is illustrated in Figure~\ref{fig:register}. Specifically, a line $s \in \mathcal{L}_{\text{src}}$ is associated with a corresponding face embedding $e_{\text{v}}(s)$ (if a active speaker is detected by TalkNet), timbre embedding $e_{\text{a}}(s)$, visual cluster label $c_{\text{v}}(s)$, and audio cluster label $c_{\text{a}}(s)$. For each visual cluster $i$, we collect its set of lines $S_{i}=\{s\in\mathcal{L}_{\text{src}}\mid c_{\text{v}}(s)=i\}$ and regard this visual cluster as a registered speaker. We then perform voting on audio clusters within $S_{i}$:

\begin{equation}
j_i=\arg\max_j \big|\{s\in S_{i}\mid c_{\text{a}}(s)=j\}\big|.
\end{equation}
The audio cluster $j_i$ receiving the highest number of votes is taken as the audio cluster corresponding to speaker $i$, and the mean timbre embedding of the lines in this cluster is used as the timbre prototype of the registered speaker:
\begin{equation}
\mu_i=\frac{1}{|T_i|}\sum_{s\in T_i} e_{\text{a}}(s),\quad T_i=\{s\in S_i\mid c_{\text{a}}(s)=j_i\}.
\end{equation}
Finally, each visual cluster $i$ is registered as a speaker, yielding $\mathcal{R}=\{(i,\mu _{i})\mid i=1,\cdots ,n_{\text{v}}\}$.  

\subsubsection{Speaker Supplementation}

Due to the audio-visual asynchrony problem of visual modality, active speaker detection cannot cover all lines in $\mathcal{L}_{\text{src}}$. For lines whose active speaker is not detected, their timbre embedding will be compared with all speaker prototype embeddings $\mu_i$ in $\mathcal{R}$ using cosine similarity, and the speaker will be labeled based on the highest similarity. However, some of these lines may contain missed speakers that TalkNet fails to detect; therefore, we design a new speaker supplementation strategy.

Specifically, we compute cosine similarity between adjacent lines in $\mathcal{L}_{\text{src}}$, and take the position of adjacent line pairs whose similarity is below threshold $\epsilon$ ($\epsilon=0.35$ in our experiments) as the boundary of groups. In this way, all lines in each group $G$ belong to the same speaker. For each line $s \in G$, we determine a new speaker score $\sigma(s)$ based on whether an active speaker is detected:  
\begin{equation}
\begin{aligned}
\sigma(s) =
\begin{cases}
1.0, & \; \; \; \text{detected} \\
\max \{\text{sim}(e_{\text{a}}(s),\mu_{i}) \mid i=1,\cdots,n_{\text{v}}\}, & \; \; \;\text{not detected}
\end{cases}
\end{aligned}
\end{equation}
We then compute an average new speaker score within group $G$ as $\sigma(G) = \sum_{s \in G} \sigma(s) / |G|$. If $\sigma(G)$ is below the threshold $\eta$ ($\eta=0.4$ in experiments), the average timbre embedding of the lines in $G$, $\mu_{G} = \sum_{s \in G} e_{\text{a}}(s)/|G|$, is registered as a new speaker. If a new speaker has already been registered, it will be merged with an existing new speaker according to the timbre embedding similarity threshold $\epsilon$. The new speaker supplementation process is shown in Table~\ref{tab:newspeaker}. See \href{https://github.com/CcQunResearch/Hermes/blob/main/Reproducibility%20Supplementary%20Material.pdf}{Supplementary Material} (1.1) for more details of SD.

\begin{table*}[!h]
\centering
\caption{The new speaker supplementation process. \textit{Similarity} represents the timbre embedding similarity of adjacent lines.}
\resizebox{0.805\textwidth}{!}{
\begin{tabular}{clcccccc}
 \Xhline{1.0pt}
 \rowcolor{gray!20}
 \textbf{Group} & \textbf{Line} & \textbf{Similarity} & \textbf{Active} & $\sigma(s)$ & $\sigma(G)$ & \textbf{Operation} \\
 \hline
 \multirow{2}{*}{1} & \texttt{I'm always with you.} & - & \ding{51} & 1.0 & \multirow{2}{*}{1.0} & \multirow{2}{*}{-}\\
 ~ & \texttt{Even when we're worlds apart.} & 0.632 & \ding{51} & 1.0 & ~ & ~ \\
 \hdashline
 \multirow{2}{*}{2} & \texttt{They want better lives,} & 0.231 & \ding{55} & 0.179 & \multirow{2}{*}{0.216} & \multirow{2}{*}{Register new speaker}\\
 ~ & \texttt{but emotion clashes with reason.} & 0.673 & \ding{55} & 0.244 & ~ \\ 
 \hdashline
 3 & \texttt{There is beauty in imperfections.} & 0.179 & \ding{51} & 1.0 & 1.0 & - \\
 \hdashline
 \multirow{2}{*}{4} & \texttt{I admired about you.} & 0.215 & \ding{55} & 0.303 & \multirow{2}{*}{0.280} & \multirow{2}{*}{\makecell[l]{Register new speaker or merge\\with existing new speakers}}\\
 ~ & \texttt{There is no prize to perfection.} & 0.731 & \ding{55} & 0.257 & ~ & ~ \\
 \Xhline{1.0pt}
\end{tabular}
}
\label{tab:newspeaker}
\end{table*}

\subsection{Terminology Identification}

\subsubsection{Overview}

Subtitles contain numerous proper nouns such as personal names, locations, and items. The accuracy of translating these terms and maintaining their consistency throughout the entire program are crucial for the viewing experience. As a specialized domain corpus, a visual media program often includes rare words or proper nouns whose meanings are highly dependent on the context. Therefore, we leverage bilingual parallel corpora and the in-context learning (ICL) capability of LLMs to construct a terminology identification training set, which is then used to train an LLM to understand and translate proper nouns in subtitles.

\subsubsection{Terminology LLM Training}

We employ an off-the-shelf LLM $\pi$ (Qwen-Max) to one-shot detect terms, their types, and translations from bilingual lines $(s, t)$ in the dataset $\mathbb{D}$. Using the speaker turns annotated by SD module as boundaries, we segment the lines into groups, where a group contains $n$ lines $s_{i}$ and their translations $t_{i}$.The input of $\pi$ is $\hat{x}\equiv \mathcal{C}\oplus \{s_i,t_i\}$, where $\mathcal{C}$ includes text such as the introduction, instructions, and one-shot examples that guide the LLM to identify terms. The output of the LLM consists of the identified term $p$ from ${s_i,t_i}$, its type $\text{tp}(p)$, and its translation $\text{tr}(p)$.

In a program, a given term may appear multiple times, with its type varying due to the randomness of LLM generation, and its translation differing because of factors such as abbreviations or aliases in the line. Therefore, for the raw candidate results identified by the LLM, $\mathbb{T}_{\text{raw}}\equiv \{(p,\text{tp}(p),\text{tr}(p))\}$, we apply filtering and voting to determine the type and translation of each term, yielding $\mathbb{T}_{\text{filter}}\equiv \{(p,\text{tp}^{\text{(c)}}(p),\text{tr}^{\text{(c)}}(p))\}$. Next, we use a Trie (prefix tree) to retrieve the terms appearing in the source language lines $\mathcal{L}_{\text{src}}$ from $\mathbb{T}_{\text{filter}}$, and then construct the dataset $\hat{\mathbb{D}}$ for training the terminology LLM $\pi _{\text{term}}$ (Qwen2.5-14B). The input of $\pi _{\text{term}}$ is $\hat{x}\equiv \mathcal{C}\oplus \{s_i\}$, and its response is the terminology retrieved from $\{s_i\}$, i.e., $\hat{y}\equiv \{(p,\text{tp}^{\text{(c)}}(p),\text{tr}^{\text{(c)}}(p))\}$. Algorithm~\ref{alg:ti} provides a detailed description of the entire training process of $\pi _{\text{term}}$. The prompt and response of the LLM are presented in \href{https://github.com/CcQunResearch/Hermes/blob/main/Reproducibility%20Supplementary%20Material.pdf}{Supplementary Material} (2.2).

\begin{algorithm}[!h]
  \caption{TI Model Training.}
  \label{alg:ti}
  \begin{algorithmic}[1]
    \Require vanilla LLM $\pi$, dataset $\mathbb{D}$.
    \Ensure terminology LLM $\pi _{\text{term}}$.
    \For {any $\hat{x}\in \mathbb{D}$} 
    \State Identify proper noun from $(s_i, t_i)$: $\pi (\hat{x})\rightarrow (p,\text{tp}(p),\text{tr}(p))$.
    \EndFor
    \State Merge results and get candidate set $\mathbb{T}_{\text{raw}}\equiv \{(p,\text{tp}(p),\text{tr}(p))\}$.
    \State Filter $\mathbb{T}_{\text{raw}}$ and vote to get $\mathbb{T}_{\text{filter}}\equiv \{(p,\text{tp}^{\text{(c)}}(p),\text{tr}^{\text{(c)}}(p))\}$.
    \For {any $s\in \mathcal{L}_{\text{src}}$} 
    \State Retrieve proper nouns for $s$ from $\mathbb{T}_{\text{filter}}$.
    \EndFor
    \State Utilize retrieval results to construct TI dataset $\hat{\mathbb{D}}\equiv \{(\hat{x},\hat{y})\}$.
    \State Train $\pi$ with dataset $\hat{\mathbb{D}}$ to obtain $\pi _{\text{term}}$. \\
    \Return $\pi _{\text{term}}$.
  \end{algorithmic}
\end{algorithm}

\begin{figure}[t]
  \centering
  \includegraphics[width=0.4\textwidth]{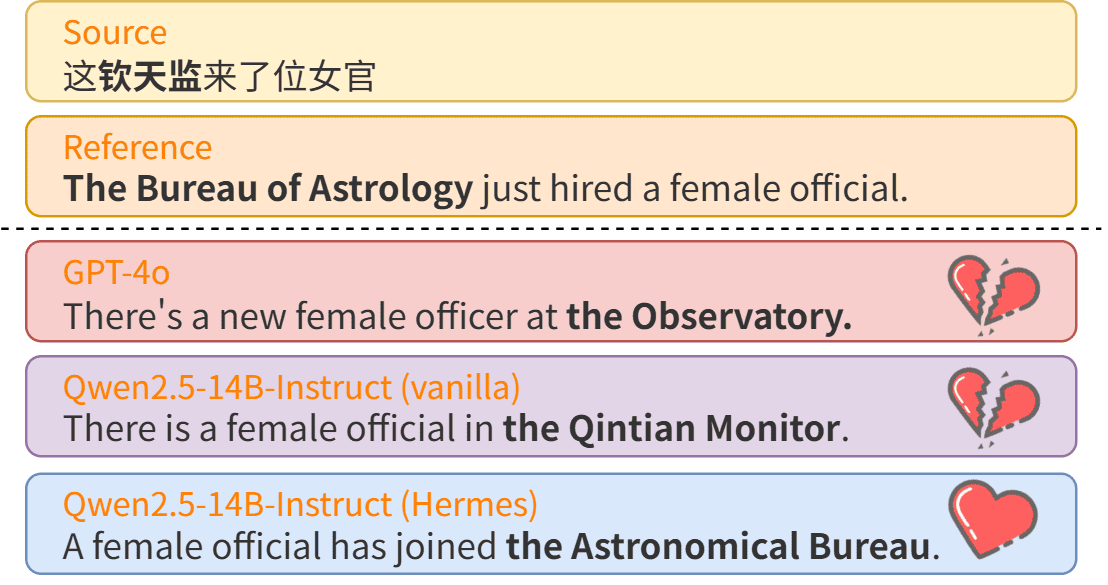}
  \caption{A \texttt{zh}$\Rightarrow$\texttt{en} terminology example. ``\begin{CJK}{UTF8}{gkai}钦天监\end{CJK}” is an ancient Chinese institution responsible for astronomical observations, calendar compilation, and weather prediction.}
  \label{fig:termexample}
\end{figure}

In practice, ICL can identify terms in subtitles with a high recall (96.9\% on a test set containing 20 programs with annotated terms), but it cannot provide suitable translations (note that only monolingual subtitles are available during inference). This is because certain rare nouns appear in special types of programs (e.g., Fantasy, Sci-Fi, Period Drama). Our TI module, which adopts a knowledge distillation-like approach, can effectively enhance the LLM's ability to translate terminology in visual media domain (as shown in Figure~\ref{fig:termexample}). See \href{https://github.com/CcQunResearch/Hermes/blob/main/Reproducibility%20Supplementary%20Material.pdf}{Supplementary Material} (1.2) for more details of TI.

\subsection{Expressiveness Enhancement}

\subsubsection{Overview}

Subtitle translation places customized requirements on translation quality. Existing studies\cite{avt2,avt3,avt4} show that the degree of literal and liberal translation in human translated corpora varies significantly across domains. Domains like visual media, literature, and religion exhibit a stronger tendency toward liberal translation compared to legislation, news, and medicine. The former emphasize the expressiveness of the translation, while the latter, containing more technical or formal texts, emphasize accuracy. Therefore, our goal is to enhance the expressiveness of translations. Previous studies \cite{translationjudge1,translationjudge2} have shown that LLMs can serve as reliable evaluators of translation accuracy. Based on the previous work \cite{sspo}, we propose SAPO to enhance the expressiveness of subtitle translation. We employ DeepSeek-V3.1 as the reward model and leverage segment-wise sampling and a fine-grained alignment loss to train an expressive translation model $\pi _{\text{st}}$.

\subsubsection{Translation Model Training}

SD and TI modules provide prior information for lines, such as character annotations and terminology translations. We then train a subtitle translation model $\pi_{\text{sft}}$ through supervised fine-tuning (SFT) on dataset $\mathbb{D}$, using the constructed $(x,y)$ pairs to fine-tune an off-the-shelf LLM. Specifically, the predictions from SD module are used to assign a speaker label $c_i$ (e.g., \texttt{[A]}) to each source line $s_i$ in $x$. In addition, we employ a vision-language model (e.g., Qwen2.5-VL-7B-Instruct) to infer the speaker’s age and gender from facial information, ensuring accurate translation of personal pronouns. To preserve the integrity of line segments, speaker turns are also used as boundaries for grouping lines. For terminology identified by TI module, we again use a Trie to retrieve the terms $(p,\text{tp}(p),\text{tr}(p))$ that appear in $s_i$ from $\mathbb{T}_{\text{filter}}$. 

\subsubsection{Sampling Strategy}

Obviously, constructing an expressive translation LLM is a preference optimization task, but we cannot directly apply general preference alignment methods (e.g., PPO \cite{po1} or DPO \cite{dpo}). This is mainly because the translation of each line depends on its context, so the input $x$ of the SFT model $\pi _{\text{sft}}$ contains multiple lines $\{s_i\}$, which requires fine-grained alignment for each line in the response $y$. We define a task that requires multi-segment local preference alignment on the LLM’s response as a \textbf{local preference optimization} task, where the reward signal is applied to segments of the response (process-supervised) rather than the entire response (outcome-supervised). SAPO provides an effective paradigm for handling such tasks.

During the training of $\pi _{\text{sft}}$, we hold out 20\% of the training corpus $\mathbb{D}$ as the SAPO training set, denoted as $\mathbb{D}_{\text{sapo}}$. In SAPO sampling phase, for an $x\in \mathbb{D}_{\text{sapo}}$ containing $n$ lines $s_{1},\dots ,s_{n}$, we sample $k$ candidate translations for each line $s_{i}$ ($k=15$ in our experiments) using $p_{i}=x,t_{1}^{\text{(c)}},\dots ,t_{i-1}^{\text{(c)}}$ as a prefix, i.e., $\pi _{\text{sft}}(t_{i}^{j}\mid p_{i}),j=1,\dots ,k$, to obtain the candidate set $\{t_{i}^{j}\mid j=1,2,\dots ,k\}$. We then deduplicate this set and add the human reference translation $t_{i}^{0}$ if it is available, resulting in $\mathcal{T}_{i}=\{t_{i}^{j}\mid j=0,1,\dots \}$. Subsequently, we use an off-the-shelf LLM (Qwen2.5-14B-Instruct in our experiments) as an evaluator $\pi _{\text{e}}$ (or as a reward model) to assess the expressiveness of $\mathcal{T}_{i}$, which yields a corresponding score sequence $\mathcal{E}_{i}$ (see Table~\ref{tab:daexample} for a demonstration and see \href{https://github.com/CcQunResearch/Hermes/blob/main/Reproducibility%20Supplementary%20Material.pdf}{Supplementary Material} (2.3) for evaluation prompt). Based on $\mathcal{E}_{i}$, we select the top-1 scoring translation $t_{i}^{\text{(c)}}$ as the prefix for the next sampling cycle. Finally, for each sample $x\in \mathbb{D}_{\text{sapo}}$, we obtain the corresponding sampling results $\mathcal{S}(x)\equiv \{(s_{i},\mathcal{T}_{i},\mathcal{E}_{i})\mid i=1,\dots ,n\}$. Algorithm~\ref{alg:sampling} details this entire process.

\begin{table}[!h]
\centering
\caption{\texttt{zh}$\Rightarrow$\texttt{en} evaluation demonstration of evaluator $\pi _{\text{e}}$.}
\resizebox{0.48\textwidth}{!}{
\begin{tabular}{lc}
 \Xhline{1.0pt}
 \rowcolor{gray!20}
 \textbf{Line} & \textbf{Score} \\
 \hline
 \begin{CJK}{UTF8}{gkai}越是光芒万丈的人，越会更快地燃尽生命的能量。\end{CJK} & - \\
 \hdashline
 \texttt{People who shine brightly will use up their life energy quickly.} & 70 \\
 \texttt{The brighter one shines, the faster one burns out.} & 90 \\
 \texttt{The more dazzling a person is, the sooner they exhaust their life’s energy.} & 85 \\
 \texttt{Those who radiate the brightest light consume their life force the fastest.} & 88 \\
 \texttt{The greater the brilliance, the swifter the flame fades.} & 92 \\
 \Xhline{1.0pt}
\end{tabular}
}
\label{tab:daexample}
\end{table}

\begin{algorithm}[!h]
  \caption{SAPO Sampling Strategy.}
  \label{alg:sampling}
  \begin{algorithmic}[1]
    \Require SFT model $\pi _{\text{sft}}$, evaluation LLM $\pi _{\text{e}}$, alignment dataset $\mathbb{D}_{\text{sapo}}$, sample number $k$.
    \Ensure sampled segment-level candidate set $\mathcal{S}(x)$.
    \For {any $x\in \mathbb{D}_{\text{sapo}}$} 
    \For {$i=1$ to $n$} 
    \For {$j=1$ to $k$} 
    \State Sample $\pi _{\text{sft}}(t_{i}^{j}\mid x,t_{1}^{\text{(c)}},\dots ,t_{i-1}^{\text{(c)}})$.
    \EndFor
    \State Deduplicate the candidate set $\{t_{i}^{j}\mid j=1,\dots ,k\}$.
    \State (Optional) Add human reference $t_{i}^{0}$ to the candidate set, get $\mathcal{T}_{i}=\{t_{i}^{j}\mid j=0,1,\dots \}$.
    \State Evaluate $\mathcal{T}_{i}$ by $\pi _{\text{e}}$, get score sequence $\mathcal{E}_{i}$.
    \State Select top-1 translation $t_{i}^{\text{(c)}}$ as the prefix of next cycle.
    \EndFor
    \EndFor \\
    \Return $\mathcal{S}(x)\equiv \{(s_{i},\mathcal{T}_{i},\mathcal{E}_{i})\mid i=1,\dots ,n\}$.
  \end{algorithmic}
\end{algorithm}

\subsubsection{Alignment Loss}

SAPO enables process-supervised alignment across multiple segments in translation LLM’s response $y$. Specifically, for all lines $\{s_i\mid i=1,\dots ,n\}$ in an $x\in \mathbb{D}_{\text{sapo}}$, we assign each $s_i$ an adaptive weight $w(s_{i})$, defined as the product of a gating function $\mathbf{1}(s_{i})$ and an importance score $\delta (s_{i})$:
\begin{equation}
w(s_{i})=\mathbf{1}(s_{i})\cdot \delta (s_{i}),
\end{equation}
where $\mathbf{1}(s_{i})$ acts as a gate to determine whether $s_i$ participates in optimization. When the sampled translations lack diversity or clear quality distinction, i.e., $|\mathcal{T}_{i}|\leq 3$ or $\max(\mathcal{E}_{i})-\min(\mathcal{E}_{i})\leq 5$, we set $\mathbf{1}(s_{i})=0$; otherwise, $\mathbf{1}(s_{i})=1$. The importance score $\delta (s_{i})$ depends on the diversity of sampled translations. Since lines with richer translations provide more potential for expressiveness enhancement, we assign higher $\delta (s_{i})$ to lines with a larger number of distinct sampled translations after deduplication, i.e., $\delta (s_{i})=|\mathcal{T}_{i}|/\sum_{j=1}^{n}|\mathcal{T}_{j}|$. We then apply the following loss function:
\begin{equation}
\mathcal{L}_{\text{sapo}}(\pi _{\theta };\pi _{\text{ref}})=-\mathbb{E}_{(x,\mathcal{S}(x))\sim \mathbb{D}_{\text{sapo}}}\left (\sum_{i=1}^{n}w(s_{i})\cdot \mathcal{L}_{\text{po}}(s_{i})\right ),
\end{equation}
where $\pi _{\theta }$ is the policy model and $\pi _{\text{ref}}$ is the reference model. SAPO is compatible with various plug-and-play preference optimization losses $\mathcal{L}_{\text{po}}$, including DPO \cite{dpo}, SimPO \cite{po17}, and GRPO \cite{grpo}. Taking DPO as an example:
\begin{equation}
\mathcal{L}_{\text{po}}(s_{i})=\text{log}\, \sigma \left (\beta \, \text{log}\frac{\pi _{\theta }(t_{i}^{\text{(c)}}\mid p_{i})}{\pi _{\text{ref}}(t_{i}^{\text{(c)}}\mid p_{i})}-\beta \, \text{log}\frac{\pi _{\theta }(t_{i}^{\text{(r)}}\mid p_{i})}{\pi _{\text{ref}}(t_{i}^{\text{(r)}}\mid p_{i})}\right ),
\end{equation}
where $\beta$ is a hyperparameter controlling the sensitivity to reward differences. 
We take the lowest-scoring translation as the rejected translation $t_{i}^{\text{(r)}}$. 

\section{Experiments}

\subsection{Experimental Settings}

We conduct experiments using the constructed dataset, which contains multi-directional subtitle corpora from the Youku platform, including \texttt{en}$\Rightarrow$\texttt{de}, \texttt{en}$\Rightarrow$\texttt{fr}, \texttt{en}$\Rightarrow$\texttt{zh}, \texttt{ko}$\Rightarrow$\texttt{zh}, \texttt{zh}$\Rightarrow$\texttt{en}, and \texttt{zh}$\Rightarrow$\texttt{th}. Each direction comprises 100–200 programs across different genres, with 10\% of the programs reserved as the test set. A detailed introduction to the dataset is provided in Appendix~\ref{sec:dc}.

We compare Hermes with the following baselines:
\begin{itemize}
\item  \textbf{VideoDubber} \cite{vd2} constructs a speech-aware, length-controlled subtitle translation model (the work most related to our proposed interlingual subtitling task).
\item  \textbf{NLLB-3.3B} \cite{nllb}, developed by Meta AI, supports translation across more than 200 languages.
\item  \textbf{MADLAD-10B} \cite{madlad}, released by Google Research, supports translation across more than 450 languages.
\item  \textbf{Google Translate} \cite{google}, a multilingual machine translation service provided by Google.
\item  \textbf{GPT-4o}, \textbf{Qwen-Max}, and \textbf{DeepSeek-V3.1}, which represent current SOTA chat LLMs.
\item  \textbf{GPT-5 Thinking} and \textbf{DeepSeek-R1}, which represent current SOTA reason LLMs. We employ the prompts in \href{https://github.com/CcQunResearch/Hermes/blob/main/Reproducibility%20Supplementary%20Material.pdf}{Supplementary Material} (2.1) to drive these frontier models for subtitle translation.
\item \textbf{Qwen2.5-14B} \cite{qwen25}, which serves as the backbone of Hermes.
\end{itemize}

Details of the experimental setup are provided in Appendix~\ref{sec:expdetail}. The source code of Hermes and our dataset is available at \url{https://github.com/CcQunResearch/Hermes}.

\subsection{Translation Quality Evaluation}

Unlike the strict emphasis on accuracy in legal and technical translation, subtitle translation prioritizes localized expression. We construct an LLM-as-a-Judge-based evaluation framework to assess translation quality, replacing traditional semantic-agnostic metrics such as BLEU. Following existing subtitle translation studies~\cite{avt2,avt3}, we focus on two primary dimensions: \textbf{Accuracy} and \textbf{Expressiveness}.
Accuracy is assessed in three aspects: 1) \textit{pronoun accuracy (PA)}, where GPT-4o extracts pronouns and their ground-truth translations from the test set to verify accuracy of the compared methods; 2) \textit{terminology consistency (TC)}, where we perform retrieval-based matching to verify whether the LLM adopts the specified terminology translations (from TI results) and evaluate the consistency ratio; and 3) \textit{translation accuracy}, evaluating whether the translation faithfully conveys the original line’s meaning. Expressiveness is evaluated in two aspects: 1) \textit{Naturalness}, assessing whether the translation is fluent and conforms to the grammar and usage of the target language or culture; and 2) \textit{Vividness}, examining whether the translation is vivid and effectively conveys the emotion, tone, and atmosphere of the original line. We employ DeepSeek-V3.1, Claude Sonnet 4, and GPT-5 as evaluators to score each subtitle segment in the test set from 0 to 100 on translation accuracy, naturalness, and vividness. Table~\ref{tab:partialeval} show the average of these models’ scores.

\begin{table*}[t]
\centering
\caption{Hermes quality evaluation. The 1st and 2nd best results are denoted as \colorbox{mlb}{\textbf{blue}} and \colorbox{mlo}{\textbf{orange}}. ST for supervised training.}
\resizebox{\textwidth}{!}{
\begin{tabular}{llccc:cc|ccc:cc|ccc:cc}
\Xhline{1.0pt}
\rowcolor{gray!20}
 ~ & ~ & \multicolumn{5}{c}{\texttt{en}$\Rightarrow$\texttt{de}} & \multicolumn{5}{c}{\texttt{en}$\Rightarrow$\texttt{fr}} & \multicolumn{5}{c}{\texttt{en}$\Rightarrow$\texttt{zh}} \\
\cline{3-17}
\rowcolor{gray!20}
 ~ & ~ & \multicolumn{3}{c}{\textbf{Accuracy}} & \multicolumn{2}{c}{\textbf{Expressiveness}} & \multicolumn{3}{c}{\textbf{Accuracy}} & \multicolumn{2}{c}{\textbf{Expressiveness}} & \multicolumn{3}{c}{\textbf{Accuracy}} & \multicolumn{2}{c}{\textbf{Expressiveness}} \\
 \cline{3-17}
 \rowcolor{gray!20}
 \multirow{-3}{*}{\textbf{Models}} & \multirow{-3}{*}{\textbf{Training}} & \textbf{PA (\%)} & \textbf{TC (\%)} & \textbf{Trans.} & \textbf{Nat} & \textbf{Vivi} & \textbf{PA (\%)} & \textbf{TC (\%)} & \textbf{Trans.} & \textbf{Nat} & \textbf{Vivi} & \textbf{PA (\%)} & \textbf{TC (\%)} & \textbf{Trans.} & \textbf{Nat} & \textbf{Vivi} \\
\hline
Gold Reference & Human & - & - & 84.8 & 83.8 & \colorbox{mlo}{\textbf{73.1}} &  - & - & 83.5 & 85.0 & 74.8 & - & - & 83.6 & 82.6 & \colorbox{mlo}{\textbf{71.5}} \\
\hdashline
VideoDubber & \multirow{4}{*}{ST} & - & - & 41.5 & 35.5 & 41.0 & - & - & 48.2 & 47.3 & 48.5 & - & - & 46.9 & 51.9 & 49.7 \\
NLLB-3.3B & ~ & - & - & 75.8 & 70.1 & 59.2 & - & - & 76.9 & 74.6 & 61.8 & - & - & 61.4 & 54.0 & 43.7 \\
MADLAD-10B & ~ & - & - & 73.2 & 65.6 & 54.5 & - & - & 73.6 & 67.8 & 57.4 & - & - & 59.7 & 55.5 & 46.3 \\
Google Translate & ~ & - & - & 89.9 & 80.6 & 62.6 & - & - & 91.9 & 84.8 & 64.3 & - & - & 84.2 & 79.7 & 54.4 \\
\hdashline
GPT-4o & \multirow{3}{*}{ICL (C)} & 70.2 & 79.3 & 94.1 & 86.7 & 66.9 & 68.4 & 85.4 & 93.2 & 88.8 & 69.5 & 82.7 & 87.3 & 89.3 & 82.3 & 59.8 \\
Qwen-Max & ~ & 72.1 & 82.1 & \colorbox{mlb}{\textbf{95.5}} & \colorbox{mlb}{\textbf{89.2}} & 68.9 & 67.3 & 88.2 & \colorbox{mlo}{\textbf{94.1}} & 89.9 & 71.6 & 86.7 & 88.4 & \colorbox{mlo}{\textbf{91.9}} & 84.4 & 61.3 \\
DeepSeek-V3.1 & ~ & 75.3 & 84.2 & 94.8 & \colorbox{mlb}{\textbf{89.2}} & 67.4 & 66.2 & 87.3 & \colorbox{mlb}{\textbf{94.9}} & \colorbox{mlb}{\textbf{90.7}} & 72.3 & 84.3 & 92.3 & 91.2 & 85.3 & 63.5 \\
\hdashline
DeepSeek-R1 & \multirow{2}{*}{ICL (R)} & 78.2 & 88.4 & 92.8 & 87.4 & 70.0 & 69.2 & 89.2 & 93.6 & 90.0 & 73.4 & 86.4 & 90.3 & 90.5 & \colorbox{mlo}{\textbf{85.7}} & 70.8 \\
GPT-5 & ~ & 79.3 & 90.1 & 93.6 & 88.6 & 72.7 & 74.3 & 91.0 & 92.2 & \colorbox{mlo}{\textbf{90.4}} & \colorbox{mlo}{\textbf{75.8}} & \colorbox{mlb}{\textbf{91.3}} & 91.2 & \colorbox{mlb}{\textbf{92.4}} & \colorbox{mlb}{\textbf{87.0}} & 71.1 \\
\hdashline
Qwen2.5-14B ($\pi_{\text{sft}}$) & SFT & \colorbox{mlb}{\textbf{84.7}} & \colorbox{mlo}{\textbf{95.4}} & 87.5 & 83.6 & 64.4 & \colorbox{mlo}{\textbf{77.2}} & \colorbox{mlo}{\textbf{93.9}} & 86.2 & 85.4 & 67.7 & \colorbox{mlo}{\textbf{89.7}} & \colorbox{mlb}{\textbf{95.3}} & 86.4 & 82.0 & 59.1 \\
Qwen2.5-14B ($\pi_{\text{st}}$) & \textbf{SAPO} & \colorbox{mlo}{\textbf{84.3}} & \colorbox{mlb}{\textbf{96.2}} & \colorbox{mlo}{\textbf{95.4}} & 88.4 & \colorbox{mlb}{\textbf{74.8}} & \colorbox{mlb}{\textbf{78.0}} & \colorbox{mlb}{\textbf{94.2}} & \colorbox{mlo}{\textbf{94.1}} & 89.2 & \colorbox{mlb}{\textbf{78.8}} & 88.2 & \colorbox{mlo}{\textbf{94.6}} & 90.6 & 84.3 & \colorbox{mlb}{\textbf{76.6}} \\
\hline
\rowcolor{gray!20}
 ~ & ~ & \multicolumn{5}{c}{\texttt{ko}$\Rightarrow$\texttt{zh}} & \multicolumn{5}{c}{\texttt{zh}$\Rightarrow$\texttt{en}} & \multicolumn{5}{c}{\texttt{zh}$\Rightarrow$\texttt{th}} \\
\cline{3-17}
\rowcolor{gray!20}
 ~ & ~ & \multicolumn{3}{c}{\textbf{Accuracy}} & \multicolumn{2}{c}{\textbf{Expressiveness}} & \multicolumn{3}{c}{\textbf{Accuracy}} & \multicolumn{2}{c}{\textbf{Expressiveness}} & \multicolumn{3}{c}{\textbf{Accuracy}} & \multicolumn{2}{c}{\textbf{Expressiveness}} \\
 \cline{3-17}
 \rowcolor{gray!20}
 \multirow{-3}{*}{\textbf{Models}} & \multirow{-3}{*}{\textbf{Training}} & \textbf{PA (\%)} & \textbf{TC (\%)} & \textbf{Trans.} & \textbf{Nat} & \textbf{Vivi} & \textbf{PA (\%)} & \textbf{TC (\%)} & \textbf{Trans.} & \textbf{Nat} & \textbf{Vivi} & \textbf{PA (\%)} & \textbf{TC (\%)} & \textbf{Trans.} & \textbf{Nat} & \textbf{Vivi} \\
\hline
Gold Reference & Human & - & - & 78.0 & 77.8 & \colorbox{mlo}{\textbf{65.8}} & - & - & 83.0 & 80.3 & 73.3 & - & - & 76.6 & 75.1 & 66.3 \\
\hdashline
VideoDubber & \multirow{4}{*}{ST} & - & - & 39.6 & 45.2 & 48.2 & - & - & 53.6 & 54.8 & 50.1 & - & - & 34.1 & 34.9 & 41.5 \\
NLLB-3.3B & ~ & - & - & 33.1 & 26.1 & 25.4 & - & - & 29.1 & 21.7 & 20.8 & - & - & 42.6 & 33.9 & 40.5 \\
MADLAD-10B & ~ & - & - & 44.9 & 42.9 & 46.7 & - & - & 45.1 & 38.9 & 37.6 & - & - & 47.9 & 50.8 & 51.0 \\
Google Translate & ~ & - & - & 54.9 & 52.8 & 52.0 & - & - & 79.8 & 66.3 & 50.2 & - & - & 55.2 & 56.2 & 54.5 \\
\hdashline
GPT-4o & \multirow{3}{*}{ICL (C)} & 76.2 & 86.3 & 80.0 & 79.9 & 58.1 & 87.1 & 90.4 & 88.5 & 83.0 & 64.6 & 73.4 & 85.3 & 88.0 & 84.4 & 67.9 \\
Qwen-Max & ~ & 72.3 & 87.3 & 83.7 & 82.5 & 61.8 & 85.3 & 91.3 & \colorbox{mlb}{\textbf{90.0}} & 85.0 & 66.8 & 72.1 & 86.2 & \colorbox{mlo}{\textbf{91.3}} & \colorbox{mlb}{\textbf{85.8}} & 69.1 \\
DeepSeek-V3.1 & ~ & 70.2 & 88.4 & 83.1 & 82.2 & 57.2 & 86.4 & 90.2 & \colorbox{mlo}{\textbf{89.5}} & 84.1 & 63.0 & 75.5 & 86.2 & 89.9 & 84.6 & 67.1 \\
\hdashline
DeepSeek-R1 & \multirow{2}{*}{ICL (R)} & 75.3 & 91.3 & 79.8 & 81.6 & 65.6 & 87.2 & 93.4 & 88.5 & 85.6 & 73.5 & 76.3 & 87.3 & 87.6 & 84.0 & 71.0 \\
GPT-5 & ~ & 75.3 & 90.2 & \colorbox{mlb}{\textbf{84.5}} & \colorbox{mlo}{\textbf{82.6}} & 65.0 & \colorbox{mlo}{\textbf{90.3}} & 92.3 & 89.1 & \colorbox{mlo}{\textbf{86.1}} & \colorbox{mlo}{\textbf{75.2}} & 80.2 & 88.4 & 88.7 & 83.9 & \colorbox{mlo}{\textbf{73.0}} \\
\hdashline
Qwen2.5-14B ($\pi_{\text{sft}}$) & SFT & \colorbox{mlb}{\textbf{81.3}} & \colorbox{mlb}{\textbf{95.9}} & 80.9 & 76.1 & 53.9 & \colorbox{mlb}{\textbf{91.2}} & \colorbox{mlo}{\textbf{95.3}} & 85.2 & 80.1 & 54.8 & \colorbox{mlo}{\textbf{84.2}} & \colorbox{mlo}{\textbf{92.7}} & 87.3 & 82.6 & 66.0 \\
Qwen2.5-14B ($\pi_{\text{st}}$) & \textbf{SAPO} & \colorbox{mlo}{\textbf{80.2}} & \colorbox{mlo}{\textbf{94.2}} & \colorbox{mlo}{\textbf{84.3}} & \colorbox{mlb}{\textbf{83.3}} & \colorbox{mlb}{\textbf{70.5}} & 88.4 & \colorbox{mlb}{\textbf{96.4}} & 88.3 & \colorbox{mlb}{\textbf{86.8}} & \colorbox{mlb}{\textbf{81.7}} & \colorbox{mlb}{\textbf{85.6}} & \colorbox{mlb}{\textbf{93.2}} & \colorbox{mlb}{\textbf{91.9}} & \colorbox{mlo}{\textbf{84.7}} & \colorbox{mlb}{\textbf{74.2}} \\
\Xhline{1.0pt}
\end{tabular}
}
\label{tab:partialeval}
\end{table*}

The results show that Hermes and frontier LLMs such as GPT-5 significantly outperform traditional translation models like MADLAD, achieving higher performance than human in terms of accuracy and naturalness. However, human translations excel in vividness, which may indicate that human translators often incorporate video context to perform more liberal translation. Reason models achieve better vividness scores compared to chat models, though their accuracy and naturalness are slightly lower in certain directions. Models trained with SAPO not only achieve substantial improvements in vividness over SFT but also show clear gains in accuracy and naturalness. Hermes achieves the highest vividness scores across all directions and occasionally surpasses frontier LLMs on other dimensions, particularly excelling across all dimensions in relatively low-resource directions such as \texttt{ko}$\Rightarrow$\texttt{zh} and \texttt{zh}$\Rightarrow$\texttt{th}. Moreover, the trained 14B model demonstrates significantly better performance than other LLMs in PA and TC. See \href{https://github.com/CcQunResearch/Hermes/blob/main/Reproducibility%20Supplementary%20Material.pdf}{Supplementary Material} (2.4) for the prompt for LLM evaluators. 

\subsection{Human Evaluation of Translation Quality}
\label{sec:humaneval}

We present the human evaluation of Hermes translation quality on \texttt{en}$\Rightarrow$\texttt{zh} and \texttt{zh}$\Rightarrow$\texttt{th}  in Table~\ref{tab:humaneval}. Considering the subjective preferences of different evaluators, we do not adopt a scoring scheme but instead conduct pairwise comparisons of translations to assess the win rate. Hermes is compared with four baselines along the dimensions of accuracy, naturalness, and vividness, as well as in a comprehensive evaluation. The human evaluation results are consistent with those in Table~\ref{tab:partialeval}, validating the reliability of our evaluation framework based on LLM-as-a-Judge. The detailed setup is provided in Appendix~\ref{sec:humanevalsetting}.

\begin{table*}[t]
\centering
\caption{Human win rate (win:tie:loss) on translation quality. The winning and losing contrasts are marked in \colorbox{mlb}{blue} and \colorbox{mlo}{orange}.}
\resizebox{\textwidth}{!}{
\begin{tabular}{clcccc|cccc}
\Xhline{1.0pt}
\rowcolor{gray!20}
~ & ~ & \multicolumn{4}{c}{\texttt{en}$\Rightarrow$\texttt{zh}} & \multicolumn{4}{c}{\texttt{zh}$\Rightarrow$\texttt{th}} \\
\cline{3-10}
\rowcolor{gray!20}
\multirow{-2}{*}{\textbf{Challenger}} & \multirow{-2}{*}{\textbf{Competitors}} & \textbf{Accuracy} & \textbf{Naturalness} & \textbf{Vividness} & \textbf{Comprehensive} & \textbf{Accuracy} & \textbf{Naturalness} & \textbf{Vividness} & \textbf{Comprehensive} \\
\hline
\multirow{4}{*}{\makecell{Hermes \\ (Qwen2.5-14B)}} & Gold Reference & \colorbox{mlb}{29:49:22} & \colorbox{mlb}{28:50:22} & \colorbox{mlb}{32:42:26} & \colorbox{mlb}{31:46:23} & \colorbox{mlb}{28:48:24} & \colorbox{mlb}{26:52:22} & \colorbox{mlb}{32:39:29} & \colorbox{mlb}{29:49:22} \\
~ & SFT Model $\pi _{\text{sft}}$ & \colorbox{mlb}{26:50:24} & \colorbox{mlb}{31:48:21} & \colorbox{mlb}{38:41:21} & \colorbox{mlb}{37:43:20} & \colorbox{mlb}{26:55:19} & \colorbox{mlb}{27:51:22} & \colorbox{mlb}{39:42:19} & \colorbox{mlb}{30:50:20} \\
~ & GPT-4o & \colorbox{mlo}{22:54:24} & \colorbox{mlo}{20:57:23} & \colorbox{mlb}{29:51:20} & \colorbox{mlb}{26:54:23} & \colorbox{mlb}{23:57:20} & \colorbox{mlo}{23:51:26} & \colorbox{mlb}{36:37:27} & \colorbox{mlb}{30:45:25} \\
~ & DeepSeek-R1 & \colorbox{mlo}{22:55:23} & \colorbox{mlo}{19:57:24} & \colorbox{mlb}{22:58:20} & \colorbox{mlo}{20:59:21} & \colorbox{mlb}{23:55:22} & \colorbox{mlo}{18:62:20} & \colorbox{mlb}{28:50:22} & \colorbox{mlb}{27:47:24} \\
\Xhline{1.0pt}
\end{tabular}
}
\label{tab:humaneval}
\end{table*}

\subsection{Speaker Diarization Evaluation}

We compare the performance of SD module in Hermes with existing speaker diarization baselines in Table~\ref{tab:sdeval}. Two audio clustering methods, VBx and spectral clustering, are adopted for experiments, and we report their audio-only clustering results. Additionally, we compare our approach with the current tri-modal SOTA method, E2CP \cite{aclsd}. The evaluation metrics include DER, JER, and Text DER. Experiments are conducted on three manually annotated test sets: 1) five Chinese programs; 2) three challenging Chinese programs (mainly dialectal shows); and 3) five English programs. Results show that Hermes’ SD module, which leverages both speech and video modalities, achieves superior overall performance, even surpassing the tri-modal E2CP method in difficult visual media scenarios. Moreover, spectral clustering outperforms VBx.

\begin{table*}[t]
\centering
\caption{Speaker diarization evaluation with baseline methods.}
\resizebox{\textwidth}{!}{
\begin{tabular}{ccccc|ccc|ccc}
\Xhline{1.0pt}
\rowcolor{gray!20}
~ & ~ & \multicolumn{3}{c}{\textbf{Chinese}} & \multicolumn{3}{c}{\textbf{Chinese-Hard}} & \multicolumn{3}{c}{\textbf{English}} \\
\cline{3-11}
\rowcolor{gray!20}
\multirow{-2}{*}{\textbf{Method}} & \multirow{-2}{*}{\textbf{Modality}} & \textbf{DER} & \textbf{JER} & \textbf{Text DER} & \textbf{DER} & \textbf{JER} & \textbf{Text DER} & \textbf{DER} & \textbf{JER} & \textbf{Text DER} \\
\hline
VBx & A & 0.13982 & 0.45223 & 0.13280 & 0.20679 & 0.57714 & 0.19021 & 0.12481 & 0.41017 & 0.11624 \\
SC & A & 0.14945 & 0.47114 & 0.14240 & 0.22255 & 0.65194 & 0.21177 & 0.20116 & 0.43402 & 0.19699 \\
E2CP & AVT & 0.13595 & 0.41524 & 0.12912 & 0.21125 & 0.58984 & 0.19982 & 0.11951 & 0.39785 & 0.11098 \\
\hdashline
Hermes (VBx) & AV & \colorbox{mlo}{\textbf{0.08528}} & \colorbox{mlo}{\textbf{0.41847}} & \colorbox{mlo}{\textbf{0.07827}} & \colorbox{mlo}{\textbf{0.10585}} & \colorbox{mlb}{\textbf{0.55233}} & \colorbox{mlo}{\textbf{0.09481}} & \colorbox{mlo}{\textbf{0.10554}} & \colorbox{mlo}{\textbf{0.31771}} & \colorbox{mlo}{\textbf{0.10258}} \\
Hermes (SC) & AV & \colorbox{mlb}{\textbf{0.08325}} & \colorbox{mlb}{\textbf{0.41444}} & \colorbox{mlb}{\textbf{0.07628}} & \colorbox{mlb}{\textbf{0.10180}} & \colorbox{mlo}{\textbf{0.55392}} & \colorbox{mlb}{\textbf{0.09039}} & \colorbox{mlb}{\textbf{0.10272}} & \colorbox{mlb}{\textbf{0.31331}} & \colorbox{mlb}{\textbf{0.09418}} \\
\Xhline{1.0pt}
\end{tabular}
}
\label{tab:sdeval}
\end{table*}

\subsection{Ablation Study}

\subsubsection{Adaptive Strategy}

We validate the impact of adaptive strategies in SAPO on model performance in Table~\ref{tab:adaptive}. The results show that the gating function $\mathbf{1}(s_{i})$ has the largest effect, indicating that the gating mechanism can significantly reduce the noise from low-diversity, irrelevant lines (e.g., simple lines such as "Good morning") during optimization. In addition, removing strategies importance score $\delta (s_{i})$ also leads to a performance degradation, highlighting the necessity of fine-grained control in the training process.

\begin{table}[!h]
\centering
\caption{Impact of adaptive strategies.}
\begin{tabular}{lccc|ccc}
\Xhline{1.0pt}
\rowcolor{gray!20}
~ & \multicolumn{3}{c}{\texttt{en}$\Rightarrow$\texttt{zh}} & \multicolumn{3}{c}{\texttt{zh}$\Rightarrow$\texttt{th}} \\
\cline{2-7}
\rowcolor{gray!20}
~ & \textbf{Acc} & \textbf{Nat} & \textbf{Vivi} & \textbf{Acc} & \textbf{Nat} & \textbf{Vivi} \\
\hline
SFT & \textit{86.5} & \textit{82.1} & \textit{59.2} & \textit{87.2} & \textit{82.4} & \textit{66.0} \\
SAPO & \textit{90.6} & \textit{84.2} & \textit{76.6} & \textit{91.9} & \textit{84.7} & \textit{74.1} \\
\hdashline
w/o $w(s_{i})$ & 88.1 & 83.2 & 67.4(↓9.2) & 89.2 & 83.4 & 70.3(↓3.8) \\ 
\quad w/o $\mathbf{1}(s_{i})$ & 89.1 & 83.4 & 70.2(↓6.4) & 89.9 & 83.9 & 71.2(↓2.9) \\
\quad w/o $\delta (s_{i})$ & 89.4 & 83.5 & 72.4(↓4.2) & 90.2 & 83.1 & 71.9(↓2.2) \\
\Xhline{1.0pt}
\end{tabular}
\label{tab:adaptive}
\end{table}

\subsubsection{Backbone Models}

Table~\ref{tab:backbone} compares the performance of models using different backbones. The results show that SAPO training leads to significant improvements across all backbones compared to the SFT model. Among them, the larger-parameter Qwen2.5-14B model achieves superior and more stable results. Due to its weaker performance in Chinese, the LLaMA-3.1-8B model performs worse than the other two models in the \texttt{en}$\Rightarrow$\texttt{zh} and \texttt{zh}$\Rightarrow$\texttt{en} directions, but still demonstrates a clear improvement over the SFT model.

\begin{table}[!h]
\centering
\caption{Impact of backbone models. The 1st and 2nd best results are marked as \colorbox{mlb}{\textbf{blue}} and \colorbox{mlo}{\textbf{orange}}.}
\resizebox{0.48\textwidth}{!}{
\begin{tabular}{llccc|ccc|ccc}
\Xhline{1.0pt}
\rowcolor{gray!20}
~ & ~ & \multicolumn{3}{c}{\texttt{en}$\Rightarrow$\texttt{de}} & \multicolumn{3}{c}{\texttt{en}$\Rightarrow$\texttt{zh}} & \multicolumn{3}{c}{\texttt{zh}$\Rightarrow$\texttt{en}} \\
\cline{3-11}
\rowcolor{gray!20}
\multirow{-2}{*}{\textbf{Method}} & \multirow{-2}{*}{\textbf{Backbone}} & \textbf{Acc} & \textbf{Nat} & \textbf{Vivi} & \textbf{Acc} & \textbf{Nat} & \textbf{Vivi} & \textbf{Acc} & \textbf{Nat} & \textbf{Vivi} \\
\hline
SFT & - & \textit{87.7} & \textit{83.4} & \textit{64.4} & \textit{86.5} & \textit{82.1} & \textit{59.2} & \textit{85.2} & \textit{80.1} & \textit{54.9} \\
\hdashline
\multirow{3}{*}{SAPO} & LLaMA-3.1-8B & \colorbox{mlo}{\textbf{94.7}} & \colorbox{mlo}{\textbf{87.2}} & \colorbox{mlo}{\textbf{73.9}} & 88.0 & 83.2 & 72.3 & 87.2 & 85.3 & 77.6\\
~ & GLM4-9B & 93.2 & 86.1 & 73.2 & \colorbox{mlo}{\textbf{88.2}} & \colorbox{mlb}{\textbf{84.4}} & \colorbox{mlo}{\textbf{74.4}} & \colorbox{mlb}{\textbf{89.2}} & \colorbox{mlo}{\textbf{85.7}} & \colorbox{mlo}{\textbf{78.2}} \\
~ & Qwen2.5-14B & \colorbox{mlb}{\textbf{95.2}} & \colorbox{mlb}{\textbf{88.3}} & \colorbox{mlb}{\textbf{74.8}} & \colorbox{mlb}{\textbf{90.6}} & \colorbox{mlo}{\textbf{84.2}} & \colorbox{mlb}{\textbf{76.6}} & \colorbox{mlo}{\textbf{88.3}} & \colorbox{mlb}{\textbf{86.8}} & \colorbox{mlb}{\textbf{81.6}} \\
\Xhline{1.0pt}
\end{tabular}
}
\label{tab:backbone}
\end{table}

\subsubsection{Alignment Loss in $\mathcal{L}_{\text{sapo}}$}

We compare the performance of SAPO under different preference alignment losses $\mathcal{L}_{\text{po}}$ in Table~\ref{tab:loss}. The results show that all losses yield substantial improvements over the SFT model, confirming the broad compatibility of SAPO. Among them, the DPO loss demonstrates consistently stable and superior performance across multiple directions. GRPO effectively leverages all sampled results rather than focusing only on preference pairs; nevertheless, it achieves the best performance only in the \texttt{en}$\Rightarrow$\texttt{zh} direction. Overall, we recommend the simple yet effective DPO loss as the primary choice for $\mathcal{L}_{\text{po}}$.

\begin{table}[!h]
\centering
\caption{Impact of $\mathcal{L}_{\text{po}}$ on translation quality.}
\resizebox{0.48\textwidth}{!}{
\begin{tabular}{llccc|ccc|ccc}
\Xhline{1.0pt}
\rowcolor{gray!20}
~ & ~ & \multicolumn{3}{c}{\texttt{en}$\Rightarrow$\texttt{de}} & \multicolumn{3}{c}{\texttt{en}$\Rightarrow$\texttt{zh}} & \multicolumn{3}{c}{\texttt{zh}$\Rightarrow$\texttt{en}} \\
\cline{3-11}
\rowcolor{gray!20}
\multirow{-2}{*}{\textbf{Method}} & \multirow{-2}{*}{$\mathcal{L}_{\text{po}}$} & \textbf{Acc} & \textbf{Nat} & \textbf{Vivi} & \textbf{Acc} & \textbf{Nat} & \textbf{Vivi} & \textbf{Acc} & \textbf{Nat} & \textbf{Vivi} \\
\hline
SFT & - & 87.7 & 83.4 & 64.4 & 86.5 & 82.1 & 59.2 & 85.2 & 80.1 & 54.9 \\
\hdashline
\multirow{3}{*}{SAPO} & DPO & \colorbox{mlb}{\textbf{95.2}} & \colorbox{mlb}{\textbf{88.3}} & \colorbox{mlb}{\textbf{74.8}} & 90.6 & \colorbox{mlo}{\textbf{84.2}} & \colorbox{mlo}{\textbf{76.6}} & \colorbox{mlb}{\textbf{88.3}} & \colorbox{mlb}{\textbf{86.8}} & \colorbox{mlb}{\textbf{81.6}} \\
~ & SimPO & 93.2 & 87.4 & 73.9 & \colorbox{mlo}{\textbf{90.7}} & 83.9 & 74.3 & 86.2 & 84.2 & 80.9 \\
~ & GRPO & \colorbox{mlo}{\textbf{94.6}} & \colorbox{mlo}{\textbf{87.9}} & \colorbox{mlo}{\textbf{74.3}} & \colorbox{mlb}{\textbf{91.2}} & \colorbox{mlb}{\textbf{85.0}} & \colorbox{mlb}{\textbf{77.1}} & \colorbox{mlo}{\textbf{87.2}} & \colorbox{mlo}{\textbf{84.9}} & \colorbox{mlo}{\textbf{81.3}} \\
\Xhline{1.0pt}
\end{tabular}
}
\label{tab:loss}
\end{table}

\subsubsection{Sampling Size}

\begin{figure}[!h]
  \centering
  \includegraphics[width=0.3\textwidth]{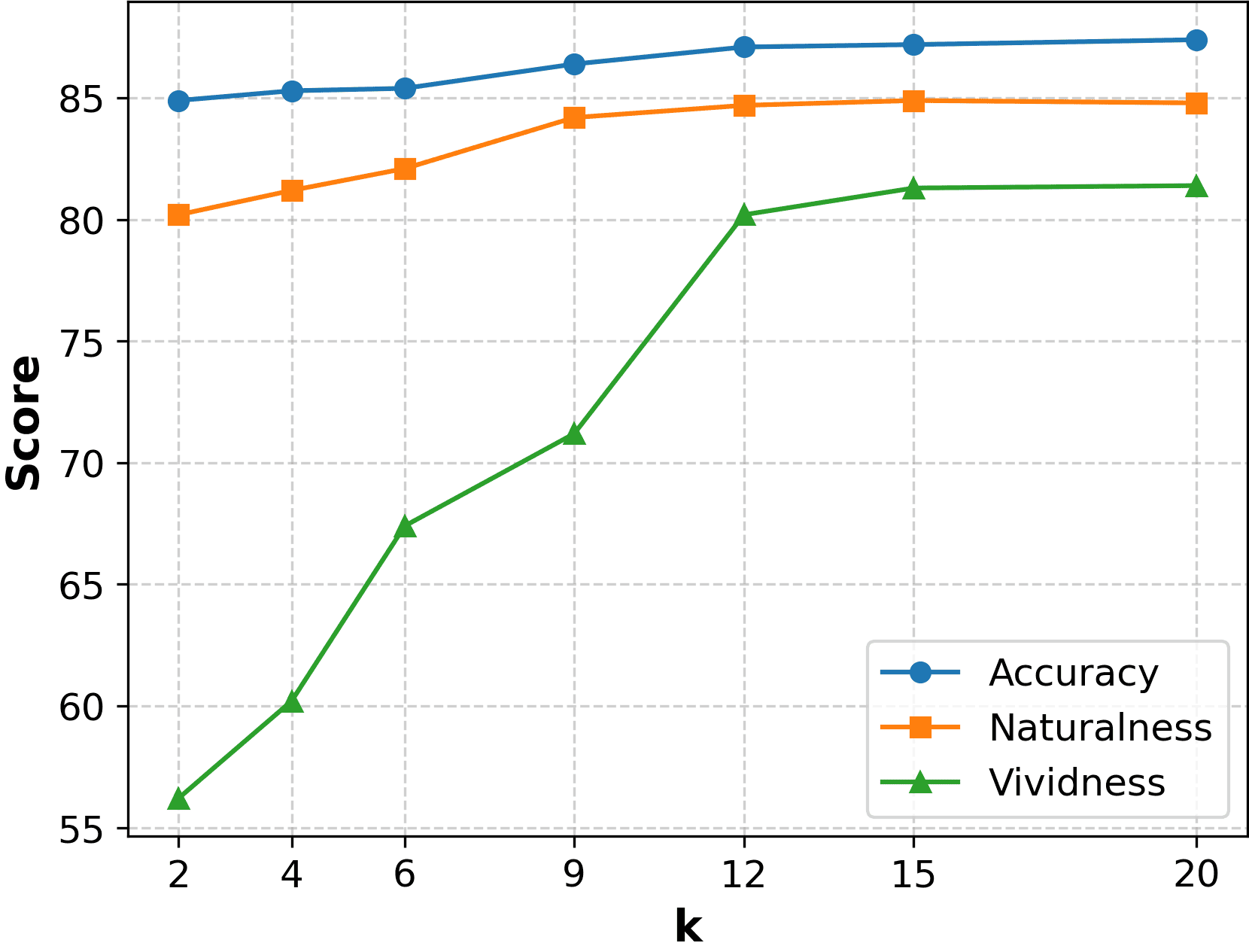}
  \caption{Impact of sampling size on translation quality.}
  \label{fig:sampingsize}
\end{figure}

The sampling size of SAPO directly affects the diversity of sampled translations. We investigate the impact of sampling size $k$ in the \texttt{zh}$\Rightarrow$\texttt{en} direction, as shown in Figure~\ref{fig:sampingsize}. The results indicate that as $k$ increases, translation quality improves across multiple dimensions, with the most pronounced gain observed in vividness, which stabilizes around $k=12$. During SAPO training, adaptive parameters such as $w(s_{i})$ and $\beta_{i}$ are computed based on the number and quality variance of sampled translations; hence, a larger sampling size leads to higher-quality training samples and more appropriate $w(s_{i})$ values. Since the SAPO loss depends on the relative quality between chosen and rejected translations, its performance tends to stabilize once the sampling size becomes sufficient. Based on these findings, we set $k=15$ in all other experiments.

\subsubsection{Model Size}

\begin{figure}[!h]
  \centering
  \includegraphics[width=0.4\textwidth]{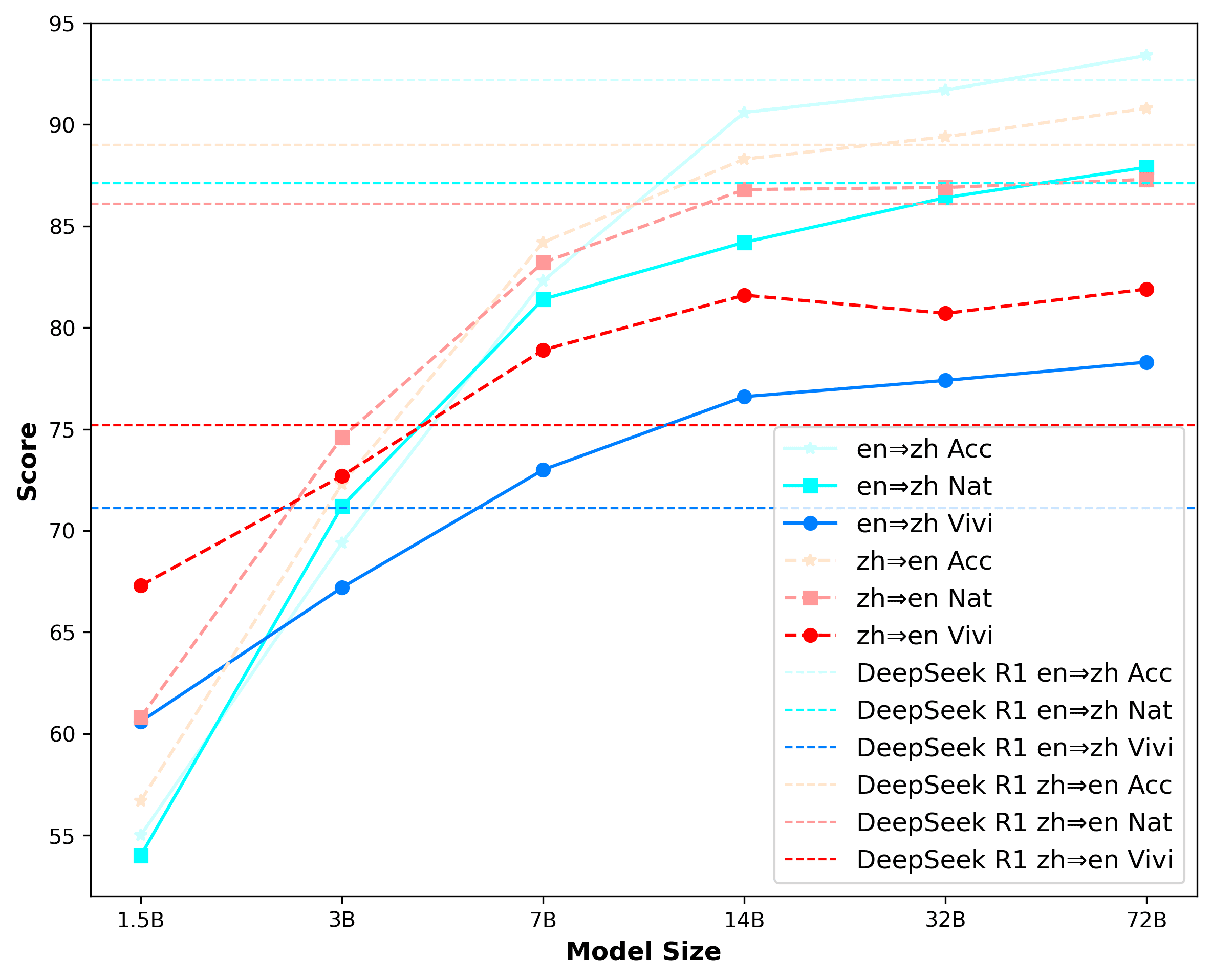}
  \caption{Impact of model size on translation quality.}
  \label{fig:modelsize}
\end{figure}

We investigate the impact of the backbone model size in SAPO in Figure~\ref{fig:modelsize}. We employ Qwen2.5 models ranging from 1.5B to 72B parameters as the base models for SAPO and evaluate their translation quality on the \texttt{en}$\Rightarrow$\texttt{zh} and \texttt{zh}$\Rightarrow$\texttt{en} directions. The results show that as the model size increases, performance improves across multiple dimensions, though the growth trend gradually plateaus. Using a 14B or 32B model generally reaches the performance level of DeepSeek-R1, further confirming the effectiveness of SAPO in improving translation quality. SAPO autonomously performs preference annotation on alignment data, enabling cost-efficient training of high-quality subtitle translation models. Overall, a 14B base model offers a balanced trade-off between cost and performance, while for subtitle translation (an offline task) larger models can be adopted when pursuing maximal performance.

\subsubsection{Data Volume}

\begin{figure}[!h]
  \centering
  \includegraphics[width=0.4\textwidth]{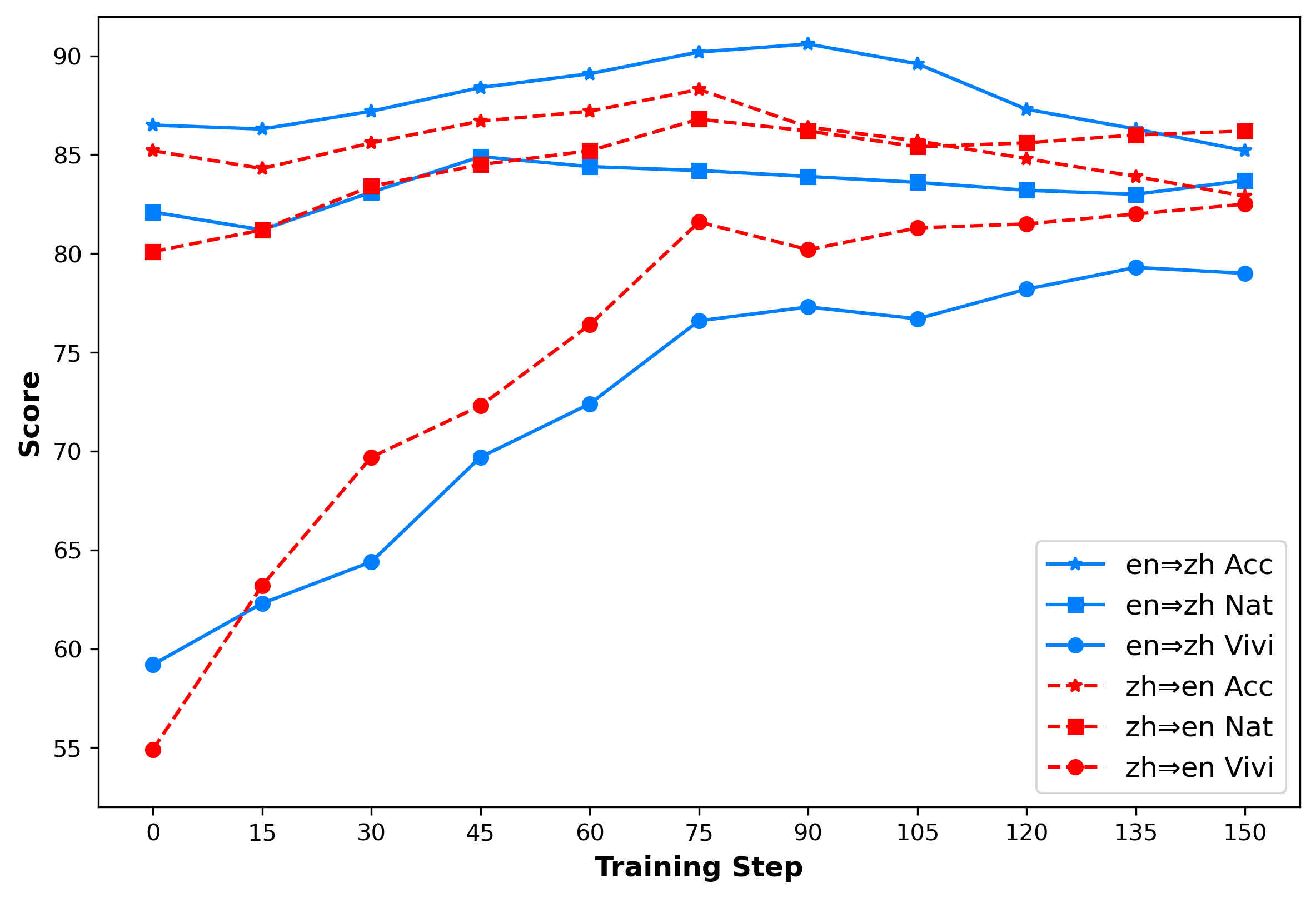}
  \caption{Impact of training data volume on performance.}
  \label{fig:step}
\end{figure}

We illustrate in Figure~\ref{fig:step} how translation quality scores vary with training steps to examine the effect of training data scale on translation quality. The results show that as the amount of training data increases, model accuracy first rises and then declines, naturalness improves slightly, while vividness increases significantly but with a diminishing growth rate. Although using more training data yields higher vividness scores, we observe evident hallucination phenomena in the model outputs, which explains the drop in accuracy. Overall, training for 75 steps proves to be the most appropriate configuration, corresponding to 7,000 prompts when the batch size is set to 96. This configuration serves as the default setting in all our experiments.

\section{Conclusion}

In this study, we propose Hermes, a unified framework for multimodal interlingual subtitling that integrates speaker diarization, terminology identification, and expressiveness enhancement to tackle coherence, consistency, and vividness challenges. Experiments on our dataset demonstrate Hermes’s superiority over existing baselines, achieving accurate, natural, and expressive subtitle translations across diverse languages.

\begin{acks}

This work is supported in part by the National Natural Science Foundation of China under Grants \#72293575 and \#62206287. 
We thank the anonymous reviewers for the valuable comments.

\end{acks}


\bibliographystyle{ACM-Reference-Format}
\bibliography{sample-base}

\appendix

\section{Related Work}

\subsection{Speaker Diarization}

Traditional speaker diarization methods predominantly rely on acoustic information, using pipelines composed of voice activity detection, speaker embedding extraction, and clustering \cite{sd1,sd2}. Although end-to-end neural diarization \cite{sd3} has improved performance, it often struggles with real-world recordings containing overlapping speech or background noise. To overcome these limitations, audio-visual approaches \cite{sd4,ava} incorporate facial and lip-motion cues to link visual activity with vocal segments, achieving better speaker attribution in multi-speaker scenes. Meanwhile, audio-textual diarization \cite{sd7} leverages linguistic or semantic information extracted from transcriptions to detect speaker turns and refine boundaries. Recent research has also explored constrained clustering \cite{sd8,sd9}, introducing weakly supervised multimodal cues into the clustering process to enhance robustness.

\subsection{LLM-as-a-Judge}

The capacity of LLMs to emulate human reasoning and evaluate specific inputs against a set of predefined rules has paved the way for "LLM-as-a-Judge" \citep{judge1,judge2}. Existing studies demonstrate that LLMs' scalability, adaptability, and cost-effectiveness render them particularly suitable for handling increasing volumes of assessment tasks that humans conventionally performed \citep{judge3,judge4}. These capabilities are crucial for deploying LLMs flexibly across diverse evaluation objectives, driving their rapid adoption in practical evaluation scenarios \citep{judge7,judge8}. Originally developed for language generation and comprehension tasks, LLMs have evolved significantly through advanced training methodologies such as Reinforcement Learning from Human Feedback (RLHF) \citep{po1}, enhancing their alignment with human values. This has allowed LLMs to transition from generative tasks to evaluative roles. Fundamentally, LLM-as-a-Judge refers to the deployment of LLMs to assess objects, actions, or decisions according to predefined rules or criteria \citep{judge10,judge11}.

\subsection{Language Model Preference Optimization}

Reinforcement learning provides an effective solution for aligning LLMs with human values and controlling text generation \citep{po6,po7}. To this end, RLHF framework based on human feedback reward models has been established \citep{po11,po13}. However, despite its effectiveness, the complexity, instability, and hyperparameter sensitivity of RLHF remain insufficiently addressed \citep{po15,po16}. Recently proposed method named DPO \citep{dpo} simplifies the RLHF framework by eliminating the need for explicit reward model construction or reinforcement learning procedures, thereby avoiding dependence on reward models. Several variants have subsequently emerged, such as SimPO, KTO, and IPO \citep{po18,po19}. Nevertheless, when applied to local preference alignment task, these methods still exhibit limitations, including coarse granularity and gradient dilution.

\section{More Details on Experimental Settings}
\label{sec:expdetail}

\subsection{Main Settings}

We primarily employ the PyTorch, vllm, and Transformers libraries to implement our approach, while utilizing DeepSpeed for multi-GPU parallel training. All experiments are conducted on a single machine equipped with eight A800 80GB SXM GPUs. The SAPO sampling phase for each translation direction takes approximately 1.5 hours, while the training phase requires about 2 hours. To ensure fairness and consistency in comparisons, we maintain identical irrelevant parameters across the main experiments, ablation studies, and extended experiments wherever possible. All critical hyperparameter settings are presented in Table~\ref{tab:hp}.

\begin{table}[h]
\centering
\caption{Hyperparameter configuration in experiments.}
\resizebox{0.48\textwidth}{!}{
\begin{tabular}{cccl}
 \Xhline{1.0pt}
 \rowcolor{gray!20}
 \textbf{Phase} & \textbf{Hyperparameter} & \textbf{Value} & \textbf{Remark} \\
 \hline
 \multirow{5}{*}{\textbf{SFT}} & n & 35 & number of lines in a prompt $x$ \\
 ~ & optimizer & AdamW & - \\
 ~ & learning rate & 1e-6 & - \\
 ~ & epoch & 4 & - \\
 ~ & batch size & 96 & - \\
 \hline
 \multirow{8}{*}{\textbf{SAPO}} & $k$ & 15 & sampling size \\
 ~ & temperature & 1.0 & \multirow{3}{*}{sampling generation hyperparameters} \\
 ~ & top k & 40 & ~ \\
 ~ & top p & 0.9 & ~ \\
 \cdashline{2-4}
 ~ & optimizer & AdamW & - \\
 ~ & learning rate & 1e-6 & - \\
 ~ & epoch & 1 & - \\
 ~ & batch size & 96 & segment-level batch size  \\
 \Xhline{1.0pt}
\end{tabular}
}
\label{tab:hp}
\end{table}

We re-implemented the VideoDubber method and utilized paid APIs from Alibaba Cloud, DeepSeek, OpenAI, and Anthropic to obtain experimental and evaluation results for models such as Qwen, DeepSeek, GPT, and Claude. When eliciting translations from these models on the test set via ICL, we employed prompt in \href{https://github.com/CcQunResearch/Hermes/blob/main/Reproducibility%20Supplementary%20Material.pdf}{Supplementary Material} (2.1), with a one-shot example to ensure structured output formatting. All LLMs used in our experiments were the latest versions as of Sep 1, 2025, specifically: qwen-max-2025-01-25\footnote{\url{https://help.aliyun.com/zh/model-studio/what-is-qwen-llm}}, DeepSeek-V3.1\footnote{\url{https://api-docs.deepseek.com/news/news250821}}, gpt-4o-2024-08-06\footnote{\url{https://platform.openai.com/docs/models}}, GPT-5\footnote{\url{https://openai.com/gpt-5}}, and Claude 4/4.1\footnote{\url{https://docs.anthropic.com/en/api/models-list}}.

For translation quality evaluation, we reserve 10\% of each dataset's programs as the test set. Due to constraints including paid API costs and inference time, we perform sampling-based evaluation rather than full-scale assessment. From these test programs, we randomly select 2,000 subtitle segments, each containing 35 lines of subtitles, thereby constructing a test set of approximately 70,000 sentences per direction. The final evaluation results represent the average scores across these 2,000 segments.

\subsection{Human Translation Quality Evaluation}
\label{sec:humanevalsetting}

We conducted human evaluation of Hermes translation quality on \texttt{en}$\Rightarrow$\texttt{zh} and \texttt{zh}$\Rightarrow$\texttt{th}. For each direction, we recruited four evaluators, all native Chinese speakers with undergraduate or master’s training in English or Thai translation. Hermes was compared against four baselines—gold reference, the SFT model $\pi _{\text{sft}}$, GPT-4o, and DeepSeek-R1—across accuracy, naturalness, and vividness, as well as through a comprehensive evaluation. The instructions provided for the comprehensive evaluation are detailed in \href{https://github.com/CcQunResearch/Hermes/blob/main/Reproducibility%20Supplementary%20Material.pdf}{Supplementary Material} (2.4), with other dimensions following a similar format. Across the two directions, the multidimensional evaluation consisted of 32 tasks, with each evaluator assigned four tasks. In each task, evaluators were given challenger and competitor translations for 400 line segments from the test set, where each segment contained 20 lines. For each segment, evaluators were asked to choose the better translation or mark them as "no significant difference". The line segments used in different tasks were sampled as disjoint subsets from the test set.

\section{Dataset Construction}
\label{sec:dc}

\subsection{Data Sources}

In this study, we conduct experiments using the dataset from the visual media program of the online video platform Youku. The dataset contains both the source language subtitles and their multilingual translated subtitles from programs on the platform. Table~\ref{tab:subtitle} presents examples of English and Chinese subtitles from the dataset (in ASS file format), where the "Start" and "End" columns indicate the start and end times of a line in the program. We use these time markers to extract the corresponding audio and video segments for each line. The translation directions covered include \texttt{en}$\Rightarrow$\texttt{de}, \texttt{en}$\Rightarrow$\texttt{fr}, \texttt{en}$\Rightarrow$\texttt{zh}, \texttt{ko}$\Rightarrow$\texttt{zh}, \texttt{zh}$\Rightarrow$\texttt{en}, and \texttt{zh}$\Rightarrow$\texttt{th}. These translation directions involve both cross language family and cross language branch pairs. As subtitle translation is a data-rich task, top online video platforms can easily collect multilingual subtitle corpora at sufficient scale. In the dataset, each translation direction contains over 100 programs, covering various genres such as films, TV series, documentaries, and animations, spanning multiple years. The statistics of the dataset are shown in Table~\ref{tab:sta}.

Unlike serious texts in domains such as legislation and medicine, which require a high degree of accuracy in translation \cite{legal1,legal2,legal3}, the dialogue in visual media programs is text that is deeply bound to the visual and speech modalities. In translation, a certain degree of accuracy loss can be tolerated; however, the translated lines must precisely convey the emotions, tone, atmosphere, cultural background, and other information present in the original lines. Therefore, human translation is not a direct literal translation of the original lines, but rather involves polishing, rewriting, and free translation to meet the localization needs of the target audience \cite{avt1,avt2}. As a result, the gold reference translations produced by human translators are polished versions based on the original text, meaning that the parallel corpus we use for training is not truly "parallel"; the original text and the translation are not fully equivalent in information (see Table~\ref{tab:subtitle} for an example). Consequently, when constructing the training set, each direction’s program is a native program in the source language.

\begin{table}[!h]
\centering
\caption{Examples of multilingual subtitles in the dataset.}
\resizebox{0.48\textwidth}{!}{
\begin{tabular}{ccl}
\Xhline{1.0pt}
\rowcolor{gray!20}
\textbf{Start} & \textbf{End} & \textbf{Text} \\
\hline
0:17:47.50 & 0:17:50.25 & \texttt{My son isn't in his right mind.} \\
0:17:51.04 & 0:17:53.95 & \texttt{His entire life, he's chased an impossible dream.} \\
0:17:59.83 & 0:18:04.20 & \texttt{But he has a good heart. Please, let him come home.} \\
0:18:04.29 & 0:18:08.66 & \texttt{A crime like this can't be overlooked.} \\
0:18:08.75 & 0:18:12.16 & \texttt{A violation of the Ethos calls for banishment,} \\
\multirow{2}{*}{0:18:12.25} & \multirow{2}{*}{0:18:16.25} & \multirow{2}{*}{\makecell[l]{\texttt{but I can sympathize with a young man's dream to} \\ \texttt{change the world.}}} \\
~ & ~ & ~ \\
\multirow{2}{*}{0:18:12.25} & \multirow{2}{*}{0:18:16.25} & \multirow{2}{*}{\makecell[l]{\texttt{Perhaps in this matter, a lesser sentence may } \\ \texttt{suffice.}}} \\
~ & ~ & ~ \\
\hdashline
0:17:47.00 & 0:17:50.25 & \begin{CJK}{UTF8}{gkai}我这儿子确实犯了不可饶恕的错误\end{CJK} \\
0:17:51.04 & 0:17:53.95 & \begin{CJK}{UTF8}{gkai}他这辈子 都在追逐一个不可能实现的梦想\end{CJK} \\
0:17:59.83 & 0:18:04.25 & \begin{CJK}{UTF8}{gkai}但他并没有坏心 求你们 放他回家吧\end{CJK} \\
0:18:04.33 & 0:18:06.70 & \begin{CJK}{UTF8}{gkai}这样严重的罪行不能轻易放过\end{CJK} \\
0:18:08.75 & 0:18:12.20 & \begin{CJK}{UTF8}{gkai}违反社会共识的人确实应该遭到驱逐\end{CJK} \\
0:18:12.29 & 0:18:16.25 & \begin{CJK}{UTF8}{gkai}但我也能体会一个年轻人梦想改变世界的雄心\end{CJK} \\
0:18:16.33 & 0:18:20.58 & \begin{CJK}{UTF8}{gkai}就这个案子来说 还是酌情予以轻判为好\end{CJK} \\
\Xhline{1.0pt}
\end{tabular}
}
\label{tab:subtitle}
\end{table}

\begin{table}[!h]
\centering
\caption{Statistics of the dataset. F, T, D, and A are abbreviations for film, TV series, documentary, and animation.}
\resizebox{0.48\textwidth}{!}{
\begin{tabular}{ccccccc}
\Xhline{1.0pt}
\rowcolor{gray!20}
\textbf{Statistic} & \texttt{en}$\Rightarrow$\texttt{de} & \texttt{en}$\Rightarrow$\texttt{fr} & \texttt{en}$\Rightarrow$\texttt{zh} & \texttt{ko}$\Rightarrow$\texttt{zh} & \texttt{zh}$\Rightarrow$\texttt{en} & \texttt{zh}$\Rightarrow$\texttt{th} \\
\hline
\textbf{cross lang. family} & \ding{55} & \ding{55} & \ding{51} & \ding{51} & \ding{51} & \ding{51} \\
\textbf{cross lang. branch} & \ding{55} & \ding{51} & \ding{51} & \ding{51} & \ding{51} & \ding{51} \\
\hdashline
\textbf{period} & 2013-2024 & 2015-2024 & 2020-2024 & 2017-2024 & 2021-2024 & 2021-2024\\
\textbf{\# programs} & 210 & 207 & 161 & 139 & 225 & 218\\
\textbf{program types} & F,T,A & F,T,A & F,T,A & T & F,T,D & F,T,D\\
\textbf{\# lines} & 1.87M & 1.72M & 1.02M & 1.54M & 1.26M & 1.21M\\
\textbf{total length (h)} & 1156 & 1033 & 852 & 802 & 501 & 479\\
\hdashline
\textbf{\# avg source token} & 7.82 & 7.67 & 7.77 & 9.87 & 6.19 & 6.08 \\
\textbf{\# avg target token} & 10.83 & 10.64 & 6.41 & 6.51 & 8.17 & 21.11 \\
\Xhline{1.0pt}
\end{tabular}
}
\label{tab:sta}
\end{table}

\subsection{Bilingual Parallel Corpus Construction}
\label{sec:saa}

Typically, the multilingual subtitles of the same program are not aligned on a sentence-by-sentence basis, primarily because the information density varies across languages, leading human translators to merge or split the original lines. In addition, subtitle files may contain noise such as scene titles or annotation information. To automatically achieve sentence-level alignment of multilingual subtitles, we design a subtitle alignment algorithm that leverages the timing of the lines.

For the source and target language line sets $\mathcal{L}_{\text{src}}$ and $\mathcal{L}_{\text{tgt}}$, we first determine their line count margin $\text{M} = \text{abs}(|\mathcal{L}_{\text{src}}| - |\mathcal{L}_{\text{tgt}}|)$, and then, for each line $s$ in $\mathcal{L}_{\text{src}}$, we search within a window of size $2\text{M}$ centered around the corresponding index in $\mathcal{L}_{\text{tgt}}$ for a line whose start time differs from that of $s$ by no more than 0.7 seconds, to serve as the corresponding translation $t$. The threshold of 0.7 seconds is chosen because, among various languages, the duration of the shortest single sentence (e.g., "Hello") is typically around 0.7 seconds. Through the above process, we can automatically obtain the dataset $\mathbb{D} \equiv \{(s, t) \in \mathcal{L}_{\text{src}} \times \mathcal{L}_{\text{tgt}}\}$ from bilingual subtitles in programs. The complexity of this process is $O(N)$.

\section{Additional Experiments}
\label{sec:furtherab}

\subsection{Speaker Turn Threshold $\epsilon$}

\begin{figure}[h]
  \centering
  \includegraphics[width=0.48\textwidth]{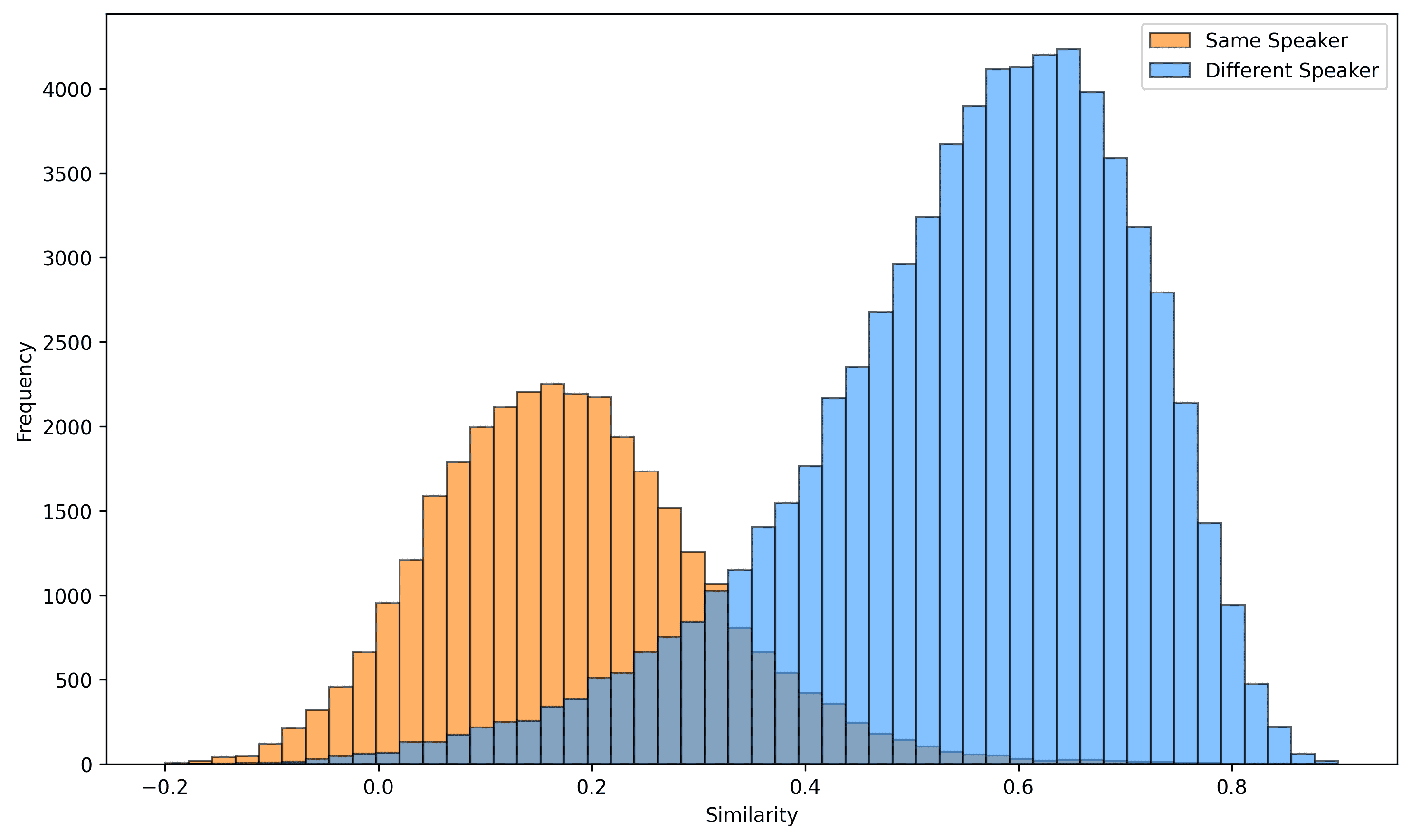}
  \caption{Label frequency distribution on Chinese program test set.}
  \label{fig:stdlabel}
\end{figure}

\begin{figure}[h]
  \centering
  \includegraphics[width=0.4\textwidth]{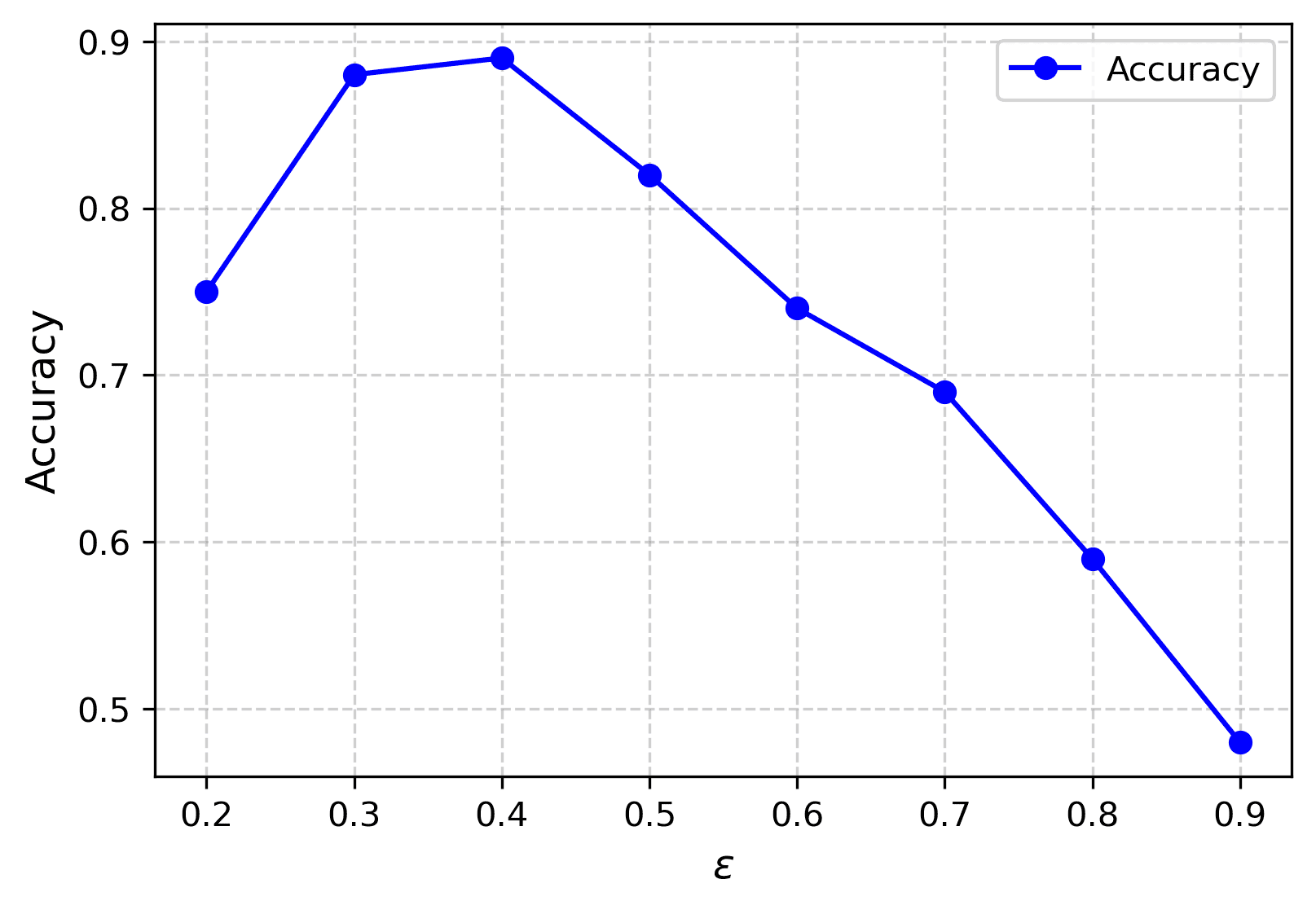}
  \caption{Impact of $\epsilon$ on speaker turn accuracy.}
  \label{fig:epsilon}
\end{figure}

The experiments were conducted on a dataset containing Chinese subtitles and audio from 10 programs, which are not included in the training data of any translation direction. The dataset comprises a total of 100,473 lines, with speaker turn labels manually annotated for each pair of adjacent lines. We extracted the timbre features of each line using ERes2NetV2 and computed the cosine similarity for every adjacent line pair. Figure~\ref{fig:stdlabel} shows the similarity-frequency distribution of the two labels on the test set.
We investigated the effect of the hyperparameter $\epsilon$ on speaker turn accuracy in Figure~\ref{fig:epsilon} to determine the optimal value. The results show that as $\epsilon$ increases, accuracy first rises and then declines. Therefore, we set $\epsilon=0.35$ as the optimal hyperparameter.

\subsection{Terminology Identification Model Size}

\begin{figure}[h]
  \centering
  \includegraphics[width=0.48\textwidth]{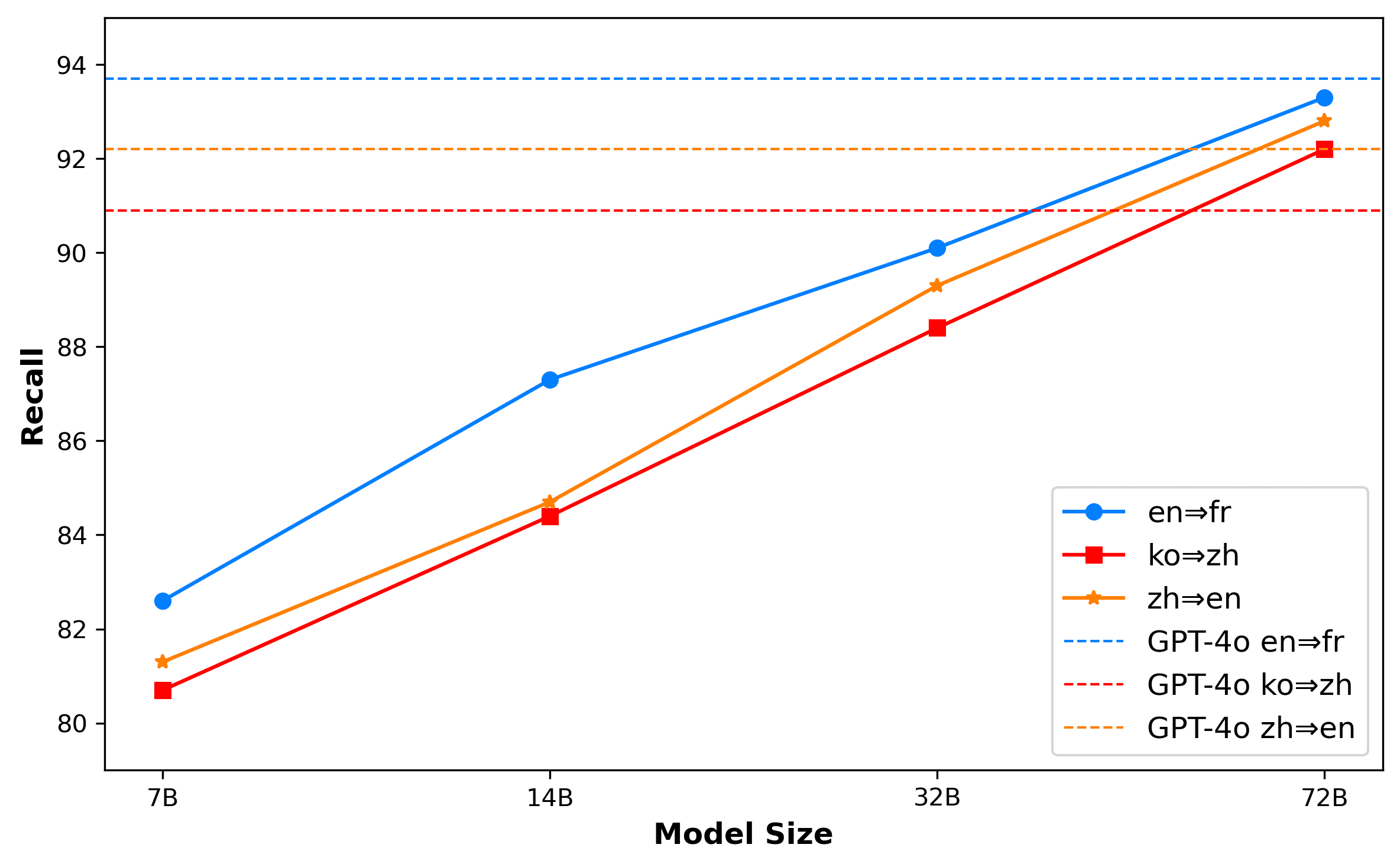}
  \caption{The impact of model size on term recall.}
  \label{fig:titmodelsize}
\end{figure}

The TIT module utilizes an off-the-shelf LLM to identify terms, their types, and translations in $\mathbb{D}$, which are then used for training $\pi_{\text{term}}$. Consequently, the recall of $\pi_{\text{term}}$ for actual terms in program source lines directly depends on the identification quality of the off-the-shelf LLM. Figure~\ref{fig:titmodelsize} illustrates the impact of different sizes of off-the-shelf LLMs on the term identification recall of $\pi_{\text{term}}$. The experimental results indicate that using larger LLMs results in a more comprehensive identification of terms within dialogues, thereby enabling $\pi_{\text{term}}$ to more effectively identify terms in the test programs. In practice, we use Qwen-Max for terminology identification.

\end{document}